\documentclass[journal]{IEEEtran} \newcommand{\isdoublecol}{1}

\usepackage[cmex10]{amsmath}
\usepackage{calc}
\usepackage{amssymb,epsfig}
\usepackage{graphicx}
\usepackage{caption}
\usepackage{cite}
\usepackage{url}
\usepackage{subcaption}
\usepackage[lined,linesnumbered,ruled]{algorithm2e}
\usepackage{color}		
\usepackage[export]{adjustbox}
\usepackage{multirow}
\usepackage{soul}
\usepackage{tabularx}
\usepackage{mathtools}

\usepackage{ifthen}

\usepackage{xr}

\usepackage{lipsum}
\usepackage{mathtools}

\usepackage{cuted}

\newcommand{\FigDir}{../TrackFacilityLocProbFigs}

\ifCLASSINFOpdf

\else

\fi

\begin{document} 

\newcommand{\lc}{\left ( }
\newcommand{\rc}{\right ) }
\newcommand{\lk}{\left \{ }
\newcommand{\rk}{\right \} }
\newcommand{\norm}[1]{\left \lVert  #1 \right \rVert}
\newcommand{\innerprod}[2]{\left \langle  {#1},  {#2} \right \rangle}
\newcommand{\floor}[1]{\left\lfloor  {#1} \right\rfloor}
\newcommand{\TM}{\texttrademark}
\newcommand{\lsq}{\left [ }
\newcommand{\rsq}{\right ] }

\title{Vehicle Tracking in Wide Area Motion Imagery via Stochastic Progressive Association Across Multiple Frames (SPAAM)}
\author{Ahmed~Elliethy,~\IEEEmembership{Student~Member,~IEEE}
        and~Gaurav~Sharma,~\IEEEmembership{Fellow,~IEEE}
\thanks{A. Elliethy is with the Department
of Electrical and Computer Engineering, University of Rochester, Rochester, NY 14627, USA (e-mail: ahmed.s.elliethy@rochester.edu).}
\thanks{G. Sharma is with the Department of Electrical and Computer Engineering, Department of Computer Science, and Department of Biostatistics and Computational Biology, University of Rochester, Rochester, NY 14627, USA (e-mail: gaurav.sharma@rochester.edu).}
\thanks{A preliminary version of part of the research presented in this paper appears in~\cite{FacilityLocationTracking:Elliethy:2017}.}
\thanks{Color  versions  of  one  or  more  of  the  figures  in  this  paper  are  available in the electronic version of this manuscript.}}


\newcommand{\IGT}{$\text{MDA-ICM}$}
\newcommand{\OT}{$\text{OT}$}
\newcommand{\pwc}[3]{{P}^{{#1}}\lc {{#2}-{#3}} \rc}
\newcommand{\pwcX}[3]{\bar{P}^{{#1}}\lc {{#2}-{#3}} \rc}
\newcommand{\Ind}{\chi}
\newcommand{\AlInd}{\boldsymbol{\Ind}} 
\newcommand{\AlIndhat}{\boldsymbol{\hat{\Ind}}} 
\newcommand{\Indi}[3]{^{#1}\chi_{{#2},{#3}}}
\newcommand{\tkA}[3]{\beta_{{#1},{#2},{#3}}}
\newcommand{\gr}{G}
\newcommand{\sgr}[2]{\mathcal{\gr}^{{#1}}_{{#2}}}
\newcommand{\sgro}[3]{\mathcal{\gr}_{{#1}\rightarrow {#2}}^{#3}}

\newcommand{\Indih}[4]{^{#1}\hat{\chi}^{#4}_{{#2},{#3}}}
\newcommand{\IndiP}[3]{^{#1}p_{{#2},{#3}}}

\newcommand{\Edg}{\varepsilon}
\newcommand{\Lt}{c}
\newcommand{\Edgi}[3]{\Edg^{#1}_{{#2},{#3}}}
\newcommand{\Edgl}[3]{\Lt^{#1}_{{#2},{#3}}}
\newcommand{\Rni}[3]{\eta^{#1}_{{#2}{#3}}}

\newcommand{\Itr}{t}
\newcommand{\Lev}{\ell}
\newcommand{\Vrs}{\mathcal{V}} 
\newcommand{\VD}[2]{z_{#2}^{#1}}
\newcommand{\VDN}[2]{v_{#2}^{#1}}
\newcommand{\Frames}{\mathcal{I}}
\newcommand{\frm}{I}
\newcommand{\VDS}{\mathcal{Z}}
\newcommand{\Edgs}{\mathcal{E}}
\newcommand{\Pths}[1]{\mathcal{P}\lc {#1} \rc}
\newcommand{\Cst}[1]{C\lc{#1}\rc}
\newcommand{\Assgn}{x}
\newcommand{\Hyps}{h}
\newcommand{\AHYPS}{\mathcal{H}}
\newcommand{\HpCst}{c}
\newcommand{\HpInd}{y}
\newcommand{\NumFrTempWn}{M}
\newcommand{\NumFrSet}{N}
\newcommand{\NumFrLev}{q}

\newcommand{\Tk}{P}
\newcommand{\Tks}{\mathcal{T}}

\newcommand{\Tkt}{\tau}
\newcommand{\Tkts}{\mathcal{L}}

\newcommand{\Hlv}{l}
\newcommand{\HlvW}{k}

\newcommand{\CstNF}{f}
\newcommand{\AssgnNF}{y}

\newcommand{\HpCstNF}{f}
\newcommand{\HpIndNF}{y}

\newcommand{\Ftr}[2]{\mathbf{f}_{#2}^{#1}}
\newcommand{\Vlc}[2]{\mathbf{v}_{#2}^{#1}}
\newcommand{\Ac}[2]{\mathbf{a}_{#2}^{#1}}

\newcommand{\Assgmnt}[4]{{Y}^{#1}_{#2}\lc #3,#4\rc}

\newcommand{\CompleteCost}[2]{{\Cst}_{#1 #2}}
\newcommand{\CompleteAssgmnt}[2]{{\Assgn}_{#1 #2}}

\newcommand{\CompleteCostNF}[2]{{\CstNF}_{#1 #2}}
\newcommand{\CompleteAssgmntNF}[2]{{\AssgnNF}_{#1 #2}}

\newcommand{\Af}{\infty}
\newcommand{\Conf}[2]{#1 \dashv #2}

\newcommand{\hyp}{h}

\newcommand{\hypC}{c}
\newcommand{\lhypC}[3]{^{#1}\hypC^{#2}_{#3}}

\newcommand{\hypS}{y}

\newcommand{\hypA}{Y}
\newcommand{\lhypA}[3]{\beta_{#1}^{{#2}\rightarrow{#3}}}
\newcommand{\lhyp}[3]{\Tk_{#1}^{{#2}\rightarrow{#3}}}
\newcommand{\lhypS}[3]{\gamma_{#1}^{{#2}\rightarrow{#3}}}
\newcommand{\lhypSOpt}[3]{\hat{\gamma}_{#1}^{{#2}\rightarrow{#3}}}
\newcommand{\LhypS}[2]{\Gamma^{{#1}\rightarrow{#2}}}
\newcommand{\LhypSOpt}[2]{\hat{\Gamma}^{{#1}\rightarrow{#2}}}

\newcommand{\GAssocK}{{X}_{\lc 1,j_{1} \rc \rightarrow \lc 2,j_{2} \rc\rightarrow\dots\rightarrow\lc {\NumFrSet},j_{\NumFrSet} \rc}}
\newcommand{\GAssocOptK}{{\hat{X}}_{\lc 1,j_{1} \rc \rightarrow \lc 2,j_{2} \rc\rightarrow\dots\rightarrow\lc {\NumFrSet},j_{\NumFrSet} \rc}}

\newcommand{\GAssocCstK}{{C}_{j_{1},j_{2}\dots j_{\NumFrSet}}}
\newcommand{\GCst}{C}

\newcommand{\GAssocSE}{{Y}_{\lc s,j_{s} \rc \rightarrow \lc s+1,j_{s+1} \rc\rightarrow\dots\rightarrow\lc e,j_{e} \rc}}
\newcommand{\GAssocSEP}[2]{{Y}_{\lc {#1},j_{{#1}} \rc \rightarrow \lc {#1}+1,j_{{#1}+1} \rc\rightarrow\dots\rightarrow\lc {#2},j_{{#2}} \rc}}
\newcommand{\GAssocSEPOPT}[2]{{\hat{Y}}_{\lc {#1},j_{{#1}} \rc \rightarrow \lc {#1}+1,j_{{#1}+1} \rc\rightarrow\dots\rightarrow\lc {#2},j_{{#2}} \rc}}

\newcommand{\GAssocCstSE}{{C}_{\lc s,j_{s} \rc \rightarrow \lc s+1,j_{s+1} \rc\rightarrow\dots\rightarrow\lc e,j_{e} \rc}}
\newcommand{\GAssocOptSE}{{\hat{Y}}_{\lc s,j_{s} \rc \rightarrow \lc s+1,j_{s+1} \rc\rightarrow\dots\rightarrow\lc e,j_{e} \rc}}
\newcommand{\LAssoc}{{X}_{\lc i,j_{i} \rc \rightarrow \lc i+1,j_{i+1}\rc}}

\newcommand{\LAssocCst}{{C}_{j_{i},j_{i+1}}}

\newcommand{\LnkG}[4]{{L}_{\lc {#1},{#2} \rc \rightarrow \lc {#3},{#4}\rc}}
\newcommand{\LAssocP}[4]{{\hat{X}}_{\lc {#1},{#2} \rc \rightarrow \lc {#3},{#4}\rc}}
\newcommand{\LAssocSg}[4]{{\hat{Y}}_{\lc {#1},{#2} \rc \rightarrow \lc {#3},{#4}\rc}}
\newcommand{\LAssocCstP}[4]{{P}_{\lc {#1},{#2} \rc \rightarrow \lc {#3},{#4}\rc}}
\newcommand{\nth}[1]{{#1}^{\text{th}}}

\newcommand{\Adj}{\mathcal{A}}
\newcommand{\DisN}{\mathcal{D}}
\newcommand{\Gsn}{\mathcal{N}}

\newcommand{\mychi}{\raisebox{0pt}[0.75ex][0.75ex]{$\chi$}}

\makeatletter
\newcommand{\thickhline}{%
    \noalign {\ifnum 0=`}\fi \hrule height 1pt
    \futurelet \reserved@a \@xhline
}
\newcolumntype{"}{@{\hskip\tabcolsep\vrule width 1pt\hskip\tabcolsep}}
\makeatother

\maketitle

\begin{abstract}
Vehicle tracking in Wide Area Motion Imagery (WAMI) relies on associating vehicle detections across multiple WAMI frames to form tracks corresponding to individual vehicles. The temporal window length, i.e., the number $\NumFrTempWn$ of sequential frames, over which associations are collectively estimated poses a trade-off between accuracy and computational complexity. A larger $\NumFrTempWn$ improves performance because the increased temporal context enables the use of motion models and allows occlusions and spurious detections to be handled better. The number of total hypotheses tracks, on the other hand, grows exponentially with increasing $\NumFrTempWn$, making larger values of $\NumFrTempWn$ computationally challenging to tackle. In this paper, we introduce SPAAM an iterative approach that progressively grows $\NumFrTempWn$ with each iteration to improve estimated tracks by exploiting the enlarged temporal context while keeping computation manageable through two novel approaches for pruning association hypotheses. First, guided by a road network, accurately co-registered to the WAMI frames, we disregard unlikely associations that do not agree with the road network. Second, as $\NumFrTempWn$ is progressively enlarged at each iteration, the related increase in association hypotheses is limited by revisiting only the subset of association possibilities rendered open by stochastically determined dis-associations for the previous iteration. The stochastic dis-association at each iteration maintains each estimated association according to an estimated probability for confidence, obtained via a probabilistic model. Associations at each iteration are then estimated globally over the $\NumFrTempWn$ frames by (approximately) solving a binary integer programming problem for selecting a set of compatible tracks. Vehicle tracking results obtained over test WAMI datasets indicate that our proposed approach provides significant performance improvements over state of the art alternatives.
\end{abstract}

\begin{IEEEkeywords}
Wide area motion imagery, vehicle tracking, association hypotheses pruning, binary integer programming, vector road network
\end{IEEEkeywords}

\IEEEpeerreviewmaketitle

\section{Introduction}
\label{sec:intro}
\IEEEPARstart{A}{erial} photography has come a long way since 1858, when the first documented aerial photograph was captured from onboard a balloon in France by Gaspard-F\'{e}lix Tournachon\cite{Hawkes:AerialPhoto:2003}. Today satellite, aircraft, drone, and balloon based aerial still-images are commonplace. Additionally, newer platforms offer motion imagery with rich  spatio-temporal information that enables a host of new applications. We focus particularly on urban area wide area motion imagery (WAMI) that offers high resolution image sequences covering large field of view (city-scale) within each frame, at temporal rates of 1-2 frames per second (fps)~\cite{WAMI_summary1,WAMI_summary2,WideAreaMotionImagerySPM:2010:Porter}. In this setting, we consider the problem of tracking the many vehicles present in the field of view. 

Tracking is an extensively researched problem (see~\cite{MotReview:2014:Luo} for recent survey). Our discussion focuses specifically on methods applicable in the WAMI setting, which poses a number of unique challenges. Typically vehicles span only few pixels and there is little spatial detail discriminating individual vehicles from each other based on appearance. Therefore, vehicles are usually represented only by their spatial locations. Moreover, the potentially large number of vehicles included within the spatial coverage of typical WAMI frames make tracking computationally demanding. The difficulty of the problem is compounded by the relatively low 1-2 fps (in contrast with typical 30 fps full motion video) temporal resolution in WAMI, due to which the spatial extent of a vehicle in adjacent frames does not typically overlap and can in fact be a fair distance apart. Further complexity and challenges are introduced by spurious and missed detections caused by imperfections in the models used for detecting vehicles against the background and due to occlusions of vehicles by over-bridges, trees, buildings, etc.

The aforementioned challenges mandate specific approaches for vehicle tracking in WAMI. Specifically, tracking is performed by first detecting the \textit{spatial locations} of vehicles (for instance, based on their movement relative to the background) and then forming vehicle tracks by associating detections that are presumed to correspond to the same vehicle over the set of WAMI frames. To obtain the associations, hypothesis tracks are first assigned costs that penalize  deviation from a motion model that typifies the common behavior of moving vehicles. Detections are then associated into tracks with the objective of minimizing the total cost for the tracks, under the constraint that a vehicle detection is assigned to at most one track. The size of the temporal context, i.e. the number of (adjoining) frames, used in formulating the motion model poses an inherent trade-off. Costs that assess the plausibility of the vehicle motion postulated by a hypothesis track are better assessed using motion models over a larger temporal context. However, the number of possible tracks increases exponentially with the number of frames, which makes it challenging to estimate the set of tracks with minimum total cost for a larger temporal context.

Existing tracking techniques handle the complexity introduced by the larger temporal context by one of  two common approaches. The first approach solves the minimum cost tracking problem exactly but with a simplified track cost that is decomposable into a sum of independent costs for pairwise associations between detections in adjacent frames that make up the track. A network is formed by arranging the detections as nodes and feasible associations between detections in adjacent frames as edges with corresponding costs. Network paths consisting of a sequence of connected edges then correspond to a feasible track whose cost is the sum of the costs of the edges in the path. The tracking problem is then posed as a minimum cost maximum flow estimation problem, which is solved exactly using one of several network flow optimization algorithms~\cite{NetworkFlows:Zhang:2008, KshortestPaths:berclaz:2011, GloballyOptimalGreedy:pirsiavash:2011, PairwiseCostsNetworkFlow:2015:Chari, FollowMeMinCostFlow:2015:Lenz}. Online tracking algorithms~\cite{VehcDetectTrack2010,DetectTrackLarge2010} that build the tracks progressively are also frequently based on the simplified cost approach. A bipartite graph is constructed for each pair of successive frames, and pairwise associations across the frames are estimated by finding the minimum cost bipartite matching. Simplified pairwise independent costs are also the basis of hierarchical track estimation methods~\cite{RobustMOTHierarchAssoc:2008:Huang, UnifiedHierarchMOT:2013:Hofmann, TrackingSportMotionModels:2013:Liu,TrackletAssociationMetricLearning:2016:Wang}. Hierarchical approaches first estimate short tracklets, then hierarchically link estimated tracklets together to form longer tracks. In~\cite{RobustMOTHierarchAssoc:2008:Huang, UnifiedHierarchMOT:2013:Hofmann, TrackingSportMotionModels:2013:Liu}, the Hungarian algorithm~\cite{hungarian:kuhn:1955} is used to estimate pairwise associations to estimate tracklets with minimum cost. In~\cite{TrackletAssociationMetricLearning:2016:Wang}, a network flow based algorithm is employed for both estimating tracklets and to link them using novel learnt tracklet discriminative metrics. In the WAMI setting, vehicle detections are only represented by their spatial locations, therefore it is beneficial to use costs based on motion models, such as constant velocity/acceleration, which rely on larger spatial context and do not meaningfully map to (independent) pairwise association costs that are required for the aforementioned algorithms. 

The second class of approaches tackles the complexity introduced by the larger temporal context by utilizing efficient methods to solve the minimum cost tracking problem approximately, without limiting the focus to independent pairwise decomposable costs. A number of iterative and/or hierarchical techniques are representative of the recent developments in this category. In~\cite{MDAHigherOrderModels:collins:2012,TensorApproxTracking:Shi:2013}, an approximate solution to the tracking problem is obtained  by iteratively re-estimating pairwise associations between two successive frames while keeping other associations fixed. In each iteration, effective ``costs'' for the pairwise associations being re-estimated are obtained by collapsing the costs over the associations that are kept fixed and the pairwise associations are re-estimated either as binary variables via the Hungarian algorithm~\cite{MDAHigherOrderModels:collins:2012} or as real-valued ``soft'' assignments computed via a rank-one tensor factorization~\cite{TensorApproxTracking:Shi:2013} which are finalized upon completion of the iterations. 
Motion models over larger temporal context can be utilized in the aforementioned approximate methods. The computational tractability obtained through the approximate approaches is, however, at a cost; the methods only ensure convergence to locally optimal solutions to the minimum cost tracking problem. Furthermore, because each iteration updates only the associations between adjacent frames often the methods converge to a weak local optima and the benefit of the cost over the larger temporal context is not fully realized. Alternative approaches, specifically focusing on a constant velocity motion model in a three frame context are proposed in~\cite{MOTUsingFrameTriplets:Butt:2013,MotLagrangianRelaxationMinCostNetworkFlow:Butt:2013}. In~\cite{MOTUsingFrameTriplets:Butt:2013}, tracklets are first estimated over a sliding three frame context using an approximate combinatorial optimization technique, consistent tracklets between adjacent three frame windows are merged and an enlarged four frame context is used to resolve conflicts; finally a min-cost network flow based global optimization is used to link together tracklets with the objective of overcoming fragmentation due to long-term occlusions. This bottom-up procedure suffers from the limitation that errors introduced in the tracklet estimation stage cannot be corrected in subsequent steps. To better maintain the global temporal context, the approach in~\cite{MotLagrangianRelaxationMinCostNetworkFlow:Butt:2013} forms an auxiliary network whose nodes represent ordered pairs of possible associations of detections from adjacent frames. The nodes are joined together by links with costs determined by the constant velocity model for each link's $3$-frame context. Because the same detection is replicated multiple times in the nodes formed by the ordered pairs, the minimum cost tracking formulation on the auxiliary network requires consistency constraints to ensure each detection is assigned only to one track. While the problem formulation does not directly correspond to the well-studied network flow problem due to these constraints, an approximate solution can be obtained by iteratively using network flow optimization techniques in combination with Lagrangian relaxation~\cite{LagrangianRelaxationBook:Reeves:1993}. The constraints are mapped to costs included in the objective function using Lagrange multipliers that are progressively adapted through the iterations to provide a final solution consistent with the constraints. The work in~\cite{HighOrderSmoothnessConstrainedGlobalTracking:Ukita:2016} considers an alternative auxiliary network based formulation where combinatorial expansion of the local context allows alternative constant velocity costs to be computed between nodes separated by a number of intervening frames. Using a shortest path algorithm iteratively in a  greedy fashion, tracks are then estimated one at time followed by elimination of nodes from the network that have been consumed by estimated tracks (which ensures that nodes are used only in at most one estimated track). The advantage of global context in these methods comes at the cost of considerable increase in complexity because of the increased size of the auxiliary network compared with the network used in the traditional network flow formulation of tracking  with decomposable costs. In particular, for the large scale of WAMI tracking problems, these approaches are not directly applicable\footnote{To solve a tracking problem over $N$ frames with $D$ detections per frame, the technique in~\cite{MotLagrangianRelaxationMinCostNetworkFlow:Butt:2013} generates an auxiliary network with $(N-1)D^2$ nodes and $(N-2)D^3$ edges compared with $N\times D$ nodes and $(N-1)D^2$ edges in the traditional flow networks.}.

In this paper, we propose SPAAM an alternative computationally-efficient, hierarchical, and iterative approach for solving the WAMI tracking problem over a relatively large temporal context without requiring track costs that are decomposable as a sum of independent pairwise association costs. In each SPAAM iteration,  the entire set of available WAMI frames are partitioned into smaller temporal windows of contiguous frames over which associations are first estimated; associations are then hierarchically linked across the windows to obtain the overall associations over the entire set of WAMI frames. The temporal windows are expanded in successive iterations to benefit from a larger temporal context. To handle the relatively large temporal windows, two innovations are proposed to limit the number of association hypotheses and to render the approach computationally tractable for the relatively large number of tracks observed in WAMI. First, guided by a pixel accurate co-registered road network (RN) with the WAMI frames, we disregard unlikely associations that do not agree with the RN; only association hypotheses that link \textit{road reachable} locations in adjacent frames are considered. Second, as the temporal window size is increased with progressing iterations, the associated increase in association hypotheses is limited by revisiting only a subset of associations. This is accomplished by: (a) stochastically preserving/dissolving associations established at the previous iteration such that the probability of preservation matches the estimated confidence probability that the association is correct, and (b) for the current iteration, maintaining the preserved associations and estimating only the associations remaining open. The confidence of every pairwise association is estimated via a probabilistic model and the estimation of open associations at each iteration is formulated as a binary integer programming (BIP) problem. In comparison with existing techniques, the proposed SPAAM approach provides two key advantages. First, because the proposed approach is not constrained to using costs that are decomposable as a sum of independent pairwise association costs, it can use more meaningful costs for the WAMI setting based on models for vehicular motion. Second, because SPAAM dissolves the (predominantly) low confidence associations and estimates remaining open associations collectively over a larger temporal context, instead of simply re-estimating the associations between an adjacent pair of frames, the method invariably overcomes the local minima problem that plagues prior methods that allowed more general costs. Results obtained with SPAAM demonstrate a significant improvement over the alternative methods benchmarked.

This paper is organized as follows. Section~\ref{sec:ProbFormulation} presents the formulation for the problem of estimating vehicle detection associations over multiple WAMI frames. Section~\ref{sec:StochProgrsvAppch} explains our proposed stochastic progressive multi-frame data association approach. Results and a comparison against alternative methods are presented in Section~\ref{sec:results}, followed by a discussion in Section~\ref{sec:discussion}. We conclude the paper in Section~\ref{sec:concl}.%

\section{Multi-frame association estimation problem formulation}
\label{sec:ProbFormulation}

\begin{table}[h]
\ifthenelse{\isdoublecol= 1}
{
\begin{tabularx}{0.49\textwidth}{l X}
}
{
\begin{tabularx}{\textwidth}{l X}
}
  $\NumFrSet$ & Total number of WAMI frames\\
  $\NumFrTempWn$ & Size of temporal window\\
  $\Frames$ & Sequence of $\NumFrSet$ WAMI frames $\frm_1,\frm_2,\dots,\frm_\NumFrSet$.\\ 
  $\mathcal{R}$ & A common coordinate system for all frames in $\Frames$\\
  $D_i$ & Number of vehicle detections (VDs) in the $\nth{i}$ WAMI frame \\
  $\VD{i}{j}$ & The location of the $j^{\text{th}}$ VD detected in the $i^{\text{th}}$ WAMI frame represented in the coordinate system $\mathcal{R}$\\
  $\gr$ & Trellis graph\\
  $\VDN{i}{j}$ & Node in the trellis graph that $\gr$ represents the VD $\VD{i}{j}$\\
  $\Vrs_{i}$ & Set of nodes in the $\nth{i}$ frame\\
  $V(\gr)$ & Set of nodes of $\gr$\\
  $\Edgi{i}{a}{b}$ & Trellis graph (directed) edge that links node $\VDN{i}{a}$ to $\VDN{i+1}{b}$\\
  $E(\gr)$ & Set of edges of $\gr$\\
  $s(\gr)$ & First frame index represented in $\gr$\\
  $e(\gr)$ & Last frame index represented in $\gr$\\
  $\Pths{\gr}$ & Set of paths that can be represented on $\gr$\\
  $\sgr{}{}$ & Edge induced sub-graph\\
  $\vartheta\lc a,b \rc$ & Minimum travel distance from $a$ to $b$ on the road network\\
  $\Tk$ & Track\\ 
  $\Adj$ & Adjacency set\\
  $\mathcal{C}\lc \VDN{i}{j} \rc$ & Set of all nodes reachable from $\VDN{i}{j}$ on the road network\\
  $\Itr$ & Iteration index\\
  $\Lev$ & Hierarchy level index\\
  $w$ & Temporal window index\\
  $k$ & Track index\\
  $(.) ^{(\Itr)}$& Value of a variable in the $\nth{\Itr}$ iteration\\
  $(.) ^{(\Itr,\Lev)}$& Value of a variable at the $\nth{\Lev}$ hierarchy level in the $\nth{\Itr}$ iteration\\
\end{tabularx}
\caption*{Key notation/symbols used in the paper}
\end{table}

\begin{figure*}[ht]
\center
\includegraphics[width=\textwidth]{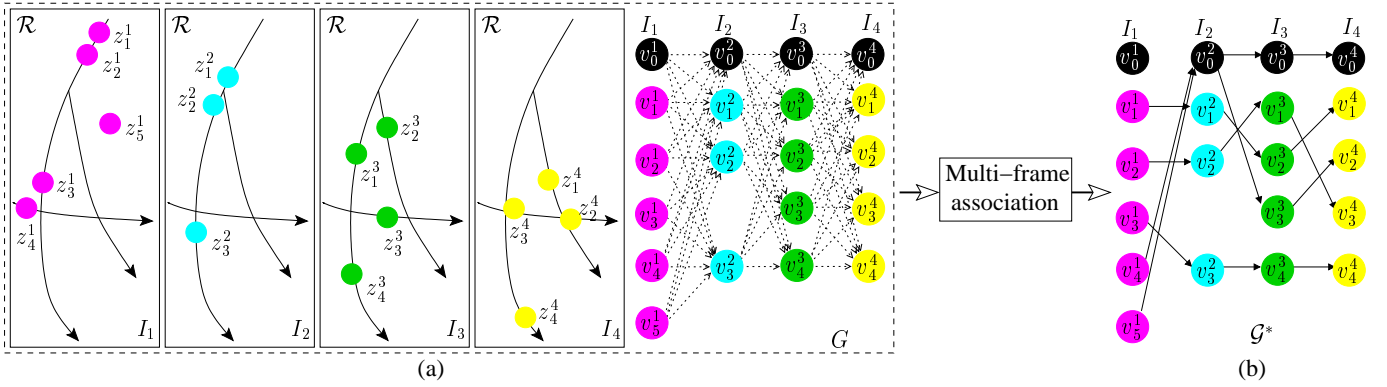}
\caption{A schematic illustration of the problem formulation via an example. The input comprises  an $\NumFrSet = 4$ frame WAMI sequence $\Frames = \lk \frm_1,\frm_2,\frm_3,\frm_4 \rk$ that is co-registered in a common reference coordinate system $\mathcal{R}$. Frame $\frm_i$ has $D_i$ VDs, with the location of the $j^{\text{th}}$ VD denoted by $\VD{i}{j}$. A trellis graph $\gr$ is obtained by associating discrete horizontal locations with the frames (in sequence) and arranging nodes vertically at the  $i^{\text{th}}$ horizontal position to represent the VDs in the corresponding frame. Node $\VDN{i}{j}$ represents the VD at location $\VD{i}{j}$, for $j=1, 2, \ldots D_i$ and a dummy node  $\VDN{i}{0}$ is introduced to handle missed/spurious detections.  The VDs in adjacent frames that can correspond to the same vehicle are connected by directed edges as shown in (a). A hypothesis vehicle track is represented on $\gr$ by a path that consists of consecutive sequence of nodes from frame $s(\gr) = 1$ to frame $e(\gr) = 4$. The goal in tracking is to estimate the best consistent subset of paths on $\Pths {\gr}$ out of the set of all possible paths, or to equivalently estimate the best edge induced sub-graph $\sgr{*}{}$ from $\gr$ such that any non-dummy node in the sub-graph must belong to exactly one path (track) as shown in (b). Note that, dummy nodes are used for missed-detections as shown in the path $\lc \VDN{1}{4},\VDN{2}{0},\VDN{3}{3}, \VDN{4}{2}\rc \in \Pths{\sgr{*}{}}$, and for identifying spurious detections as shown in $\lc \VDN{1}{5},\VDN{2}{0},\VDN{3}{0}, \VDN{4}{0}\rc \in \Pths{\sgr{*}{}}$.}
\label{Fig:ProblemDef}
\end{figure*}

Consider the problem of tracking individual vehicles over a set $\Frames = \{\frm_i\}_{i=1}^{\NumFrSet}$ of $\NumFrSet$ WAMI frames, where the WAMI frames in $\Frames$ are co-registered with a vector road network defined in a common reference\footnote{Specifically, we use a 2D Cartesian map coordinate system obtained via azimuthal orthographic map projection (AOMP)~\cite{MapProj:snyder:1987} of the spherical latitude and longitude coordinates.} coordinate system $\mathcal{R}$ (for example using~\cite{RoadNetRegistration:tip:2016,Elliethy:RegAerialI2VecRdmp:EI2016}). We adopt a tracking by detection paradigm: vehicle locations are detected in the individual frames and the goal is to associate the detections of a vehicle over the entire set of WAMI frames to form a {\em track} per vehicle. Frame $I_i$ contains $D_i$ vehicle detections (VDs) and $\VD{i}{j} = (x^i_j,y^i_j)$ denotes the spatial location of the $j^{\text{th}}$ VD in the $i^{\text{th}}$ frame in the coordinate system $\mathcal{R}$. 

To describe the problem and the solution approach, we use a 2D trellis graph  $\gr$ constructed as follows for a temporal window of WAMI frames, whose starting and ending frame numbers we denote by $s(\gr)$ and $e(\gr)$, respectively. Along the horizontal axis, nodes in the trellis graph $\gr$ lie along $T(\gr) = \lc e(\gr) - s(\gr)  \rc + 1$ discrete positions  corresponding to the frame instances from $s(\gr)$ through $e(\gr)$ arranged in increasing order from left to right. At the horizontal position for frame $i$, along the vertical axis, $D_i + 1$ distinct nodes $\Vrs_i = \lk \VDN{i}{0},\VDN{i}{1},\dots,\VDN{i}{D_i} \rk$  are placed. For $j=1, 2, \ldots D_i$, the node  $\VDN{i}{j}$ represents the  $j^{\text{th}}$ VD in the $i^{\text{th}}$ frame (at spatial location $\VD{i}{j}$ in $\mathcal{R}$) and $\VDN{i}{0}$
represents a dummy node that is introduced to account for missed-detections and for identifying spurious detections (as described later). Edges are introduced in the trellis graph linking nodes across adjacent frames\footnote{The trellis graph is characterized by the property that the only permitted edges are those that link nodes across adjacent temporal locations.} that can correspond to detections of the same vehicle, or that link a VD to a dummy node. We denote by $\Edgi{i}{a}{b}$ the (directed) edge that links $\VDN{i}{a}$ with $\VDN{i+1}{b}$  and by $\Adj \lc \VDN{i}{a} \rc$ the adjacency set for $\VDN{i}{a}$ that contains all VD nodes in the $\nth{(i+1)}$ frame linked with $\VDN{i}{a}$. In the absence of additional information, $\Adj \lc \VDN{i}{a} \rc$ contains all other nodes in the $\nth{(i+1)}$ frame. In practice, however, physical considerations rule out a number of possibilities and the adjacency set  $\Adj \lc \VDN{i}{a} \rc$ is much smaller than the full set of nodes in the $\nth{(i+1)}$ frame. For example, no edges are permissible between VD nodes whose spatial locations are further apart than the maximal distance that a vehicle can traverse in the inter-frame interval. The sets  $V\lc \gr\rc = \bigcup_{i=s(\gr)}^{e(\gr)} \Vrs_i$  and $E\lc \gr\rc = \lk \Edgi{i}{a}{b} \left| \VDN{i+1}{b}  \in \Adj \lc \VDN{i}{a} \rc, \VDN{i}{a}\in \Vrs_i, \VDN{i+1}{b}\in \Vrs_{i+1} \right.\rk$ represent the complete set of nodes and edges, respectively, in the graph $\gr$. An example trellis graph for a temporal window starting at frame  $s(\gr)=1$ and ending at frame  $e(\gr)=4$ is shown in Fig.~\ref{Fig:ProblemDef} (a). 

Within the temporal support of $\gr$, a vehicle track is represented as a sequence $\Tk_k= \lc \VDN{s(\gr)}{\zeta^{k}_{s(\gr)}}, \VDN{s(\gr)+1}{\zeta^{k}_{s(\gr)+1}}, \dots, \VDN{e(\gr)}{\zeta^{k}_{e(\gr)}}\rc$ of nodes $\zeta^{k}_{i} \in \lk 0,1,\dots,D_i \rk$ corresponding to the locations of a (single) vehicle in the frames. Specifically, $\zeta^{k}_{i} \in \Tk_k$ implies that the vehicle track passes through the location $\VD{i}{j}$ if $j \neq 0$ and that the vehicle is not detected in the $i^{\text{th}}$  frame if $j = 0$. Alternatively, to simplify notation, we also denote the track as $\Tk_k= \lc \VDN{s(\gr)}{{k}_{s(\gr)}},\VDN{s(\gr)+1}{{k}_{s(\gr)+1}},\dots, \VDN{e(\gr)}{{k}_{e(\gr)}}\rc$  at the cost of some precision. 

Because only vehicle locations are available in our setting, for estimating tracks, we proceed as follows. We denote by $\Pths {\gr}$ the set of all feasible tracks on $\gr$ and for each feasible track $\Tk_k \in \Pths {\gr}$, we assign a cost $C\lc \Tk_k \rc$ that assesses the plausibility of the vehicle movement predicated by the track. The cost decreases monotonically as the corresponding vehicle movement becomes more plausible. The objective of our tracking problem is then formulated as the task of estimating the best, i.e. minimum total cost, consistent subset of tracks from $\Pths {\gr}$, where ```consistent'' means that every non-dummy node belongs to exactly one track. In the graph formulation, the problem of estimating the best consistent subset of tracks is equivalent to the problem of selecting the minimum-cost feasible edge-induced sub-graph $\sgr{*}{}$ from the set  $\mathcal{F}\lc \gr \rc$ of all feasible edge-induced sub-graphs of $\gr$. A feasible edge-induced sub-graph  $\sgr{}{} \in \mathcal{F}\lc \gr \rc$ is a graph that has the same set of nodes of $\gr$, i.e. $V\lc \sgr{}{} \rc = V\lc \gr \rc$, but has a subset of edges from $\gr$ such that any non-dummy node in $V\lc\sgr{}{}\rc$ belongs to exactly one track in $\sgr{}{}$ as illustrated in the example shown in Fig.~\ref{Fig:ProblemDef} (b). Formally, a feasible edge-induced sub-graph $\sgr{}{} \in \mathcal{F}\lc \gr \rc$ is defined by the constraints
\begin{equation}
\begin{aligned}
& \sum\limits_{\Tk_k \in \Pths {\gr}}^{} \mychi_{\Pths {\sgr{}{}}}(\Tk_k) \; \mychi_{\Tk_k}(\VDN{i}{j}) = 1,\; \forall \VDN{i}{j} \in V\lc \sgr{}{} \rc, j\neq 0,\\
& V\lc \sgr{}{} \rc = V\lc \gr \rc,
\end{aligned}
\label{eq:ILPConstraint1}
\end{equation}
where, for any set $A$,  $\mychi_{A}(\cdot)$ is the indicator function, defined as
\begin{equation}
\mychi_{A}(x):= \begin{cases}
							1&{\text{if }}x\in A,\\
							0&{\text{if }}x\notin A.
							\end{cases}
\end{equation}
Our tracking problem is then formulated as
\begin{equation}
\sgr{*}{} = \arg\min\limits_{\sgr{}{} \in \mathcal{F}\lc \gr \rc } \psi \lc \sgr{}{} \rc,
\label{eq:OverallOptmz}
\end{equation}
where
\begin{align}
\psi \lc \sgr{}{} \rc = &\sum\limits_{\Tk_k \in \Pths {\sgr{}{}}}  C \lc \Tk_k \rc.
\label{eq:Psi}
\end{align}

\begin{figure}[ht]
\center
\ifthenelse{\isdoublecol= 1}
{
\includegraphics[width=0.49\textwidth]{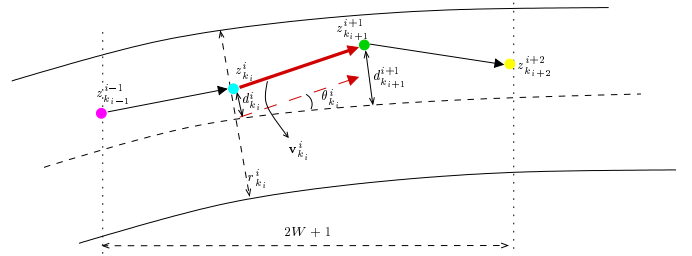}
}
{
		\resizebox{0.8\textwidth}{!}{
         \input{\FigDir/Features.pstex_t}
        }
}
         \caption{Schematic illustrating the metrics used for modeling the cost of a track $\Tk_k$.}
         \label{Fig:Features}
\end{figure}

We model the global cost $C$ of a track $\Tk_k$ as exponentiated weighted sum of metrics that penalize (a) motion irregularities, (b) road network dis-agreement, and (c) number of dummy nodes along the track. The metrics comprise (non-independent) pairwise association costs estimated over the local context associated with every pairwise association $\Edgi{i}{k_i}{k_{i+1}}$ in the track $\Tk_k$. Specifically, the cost of the track $\Tk_k$ is defined as in \eqref{eq:HypsCost}, 
\ifthenelse{\isdoublecol= 1}
{
\begin{figure*}[ht]
\normalsize
\begin{equation}
C\lc \Tk_k \rc = - \exp \lc - \sum\limits_{i=s(\gr)}^{e(\gr)} \frac{\sigma_m \Gamma \lc \Edgi{i}{k_i}{k_{i+1}} \rc +  \sigma_d R_d \lc \Edgi{i}{k_i}{k_{i+1}} \rc + \sigma_{\theta} R_{\theta} \lc \Edgi{i}{k_i}{k_{i+1}} \rc}{L_p \lc \Tk_k \rc}  + \frac{\sigma_g  L_e \lc \Edgi{i}{k_i}{k_{i+1}} \rc}{T(\gr)} \rc ,
\label{eq:HypsCost}
\end{equation}
\hrulefill
\vspace*{1pt}
\end{figure*}
}
{
\begin{align}
C\lc \Tk_k \rc &= - \exp \lc - \sum\limits_{i=s(\gr)}^{e(\gr)} \frac{\sigma_m \Gamma \lc \Edgi{i}{k_i}{k_{i+1}} \rc +  \sigma_d R_d \lc \Edgi{i}{k_i}{k_{i+1}} \rc + \sigma_{\theta} R_{\theta} \lc \Edgi{i}{k_i}{k_{i+1}} \rc}{L_p \lc \Tk_k \rc}  + \frac{\sigma_g  L_e \lc \Edgi{i}{k_i}{k_{i+1}} \rc}{T(\gr)} \rc ,
\label{eq:HypsCost}
\end{align}
}
where:
\begin{itemize}
\item  $\sigma_m$, $\sigma_d$, $\sigma_{\theta}$, and $\sigma_g$ are nonnegative weighting factors, 
\item $L_p \lc \Tk_k \rc$ is the length of $\Tk_k$ excluding dummy nodes, i.e., $L_p \lc \Tk_k \rc = | \{ \VDN{i}{k_{i}} | \VDN{i}{k_{i}} \in \Tk_k,\; k_{i} \neq 0 \} |$,
\item $\Gamma \lc \Edgi{i}{k_i}{k_{i+1}} \rc$ is a motion irregularity penalty defined as
\begin{equation}
\Gamma \lc \Edgi{i}{k_i}{k_{i+1}} \rc = 1-\frac{ \Gamma_m \lc \Edgi{i}{k_i}{k_{i+1}} \rc + \Gamma_\theta \lc \Edgi{i}{k_i}{k_{i+1}} \rc }{2} ,
\end{equation}
where $\Gamma_m \lc \Edgi{i}{k_i}{k_{i+1}} \rc$ and $\Gamma_\theta \lc \Edgi{i}{k_i}{k_{i+1}} \rc$ are normalized measures (in the rage $[0,1]$) that quantify the similarity, in magnitude and direction, respectively, of the velocity for the $\nth{i}$ frame with the velocities for other frames in a temporal window of $(2W+1)$ frames centered about the $\nth{i}$ frame. The computation of these terms
is illustrated using Fig.~\ref{Fig:Features}. The velocity at the $\nth{j}$ frame is computed as  $\Vlc{j}{k_{j}} = \lc \VD{j+1}{k_{j+1}} - \VD{j}{k_{j}} \rc / \Delta t$ where $\Delta t$ is the inter-frame duration, and inspired by ~\cite{TensorApproxTracking:Shi:2013}, the magnitude and direction similarity terms are obtained as
\begin{align}
\Gamma_m \lc \Edgi{i}{k_i}{k_{i+1}} \rc & = \frac{1}{2W} \sum\limits_{\substack{m=-W,\\ W\neq 0}}^{W} \frac{2 \norm{\Vlc{i}{k_{i}}} \norm{\Vlc{i-m}{k_{i-m}}}}{\norm{\Vlc{i}{k_{i}}}^2 +\norm{\Vlc{i-m}{k_{i-m}}}^2}, \\
\Gamma_\theta \lc \Edgi{i}{k_i}{k_{i+1}} \rc & = \frac{1}{2W} \sum\limits_{\substack{m=-W,\\ W\neq 0}}^{W} \frac{\innerprod{\Vlc{i}{k_{i}}}{\Vlc{i-m}{k_{i-m}}}}{ \norm{\Vlc{i}{k_{i}}} \norm{\Vlc{i-m}{k_{i-m}}}}.
\end{align}
The term $\Gamma_\theta \lc \Edgi{i}{k_i}{k_{i+1}} \rc$ represents the average of cosines of the angles between the direction of the velocity vector $\Vlc{i}{k_{i}}$ and the directions of the other velocity vectors in the temporal window. If the motion of the track is smooth within the temporal window, both $\Gamma_m \lc \Edgi{i}{k_i}{k_{i+1}} \rc$ and $\Gamma_\theta \lc \Edgi{i}{k_i}{k_{i+1}} \rc$ are close to one and $\Gamma \lc \Edgi{i}{k_i}{k_{i+1}} \rc$ is close to zero. For tracks with dummy nodes, velocities are computed between actual VD locations that become adjacent once the dummy nodes are skipped.

\item $R_d \lc \Edgi{i}{k_i}{k_{i+1}} \rc$ and $R_\theta \lc \Edgi{i}{k_i}{k_{i+1}} \rc$ are costs penalizing, respectively, the distance and direction disagreement with the  road network, computed using the factors illustrated in Fig.~\ref{Fig:Features}. Specifically, $R_d \lc \Edgi{i}{k_i}{k_{i+1}} \rc = d^{i}_{k_i} / r^{i}_{k_i}$ is the distance $d^{i}_{k_i}$ of the VD location $\VD{i}{k_i}$ to the nearest point on the center line of the road, normalized through division by the factor $r^{i}_{k_i}$ that represents the number of lanes on that road, and $R_\theta \lc \Edgi{i}{k_i}{k_{i+1}} \rc = 1-\cos \lc \theta^{i}_{k_i}\rc$, where $\theta^{i}_{k_i}$ is the angle between the velocity vector $\Vlc{j}{k_{j}}$ and the road (in the direction of travel) at the nearest point on the center line of the road. The terms $R_d \lc \Edgi{i}{k_i}{k_{i+1}} \rc$ and $R_\theta \lc \Edgi{i}{k_i}{k_{i+1}} \rc$ are set to zero when $\VDN{i}{k_i}$ represents a dummy node.

\item $L_e \lc \Edgi{i}{k_i}{k_{i+1}} \rc =  \mychi_{\lk 0 \rk}(k_{i+1})$ indicates whether $ \VDN{i+1}{k_{i+1}}$ is a dummy node or not.

\end{itemize}

The exponentiation in~\eqref{eq:HypsCost} moderates the cost for each component and the overall cost for a hypothesis track is in the interval $[-1, 0)$. Also, the form of~\eqref{eq:HypsCost} ensures that tracks rated poorly for any one of the penalty terms also have an overall low cost.

\section{Stochastic progressive association across multi-frames (SPAAM)}
\label{sec:StochProgrsvAppch}

To track the vehicles in the WAMI image sequence $\Frames$, we wish to solve~\eqref{eq:OverallOptmz}. For our proposed cost and typical number of frames, this multi-frame association problem is computationally intractable. In particular, the constrained minimization problem in~\eqref{eq:OverallOptmz} is an instance of BIP, which is $NP$-\textit{complete} in its general form~\cite{CombinatOptimizationComplexity:Papadimitriou:2013}. Although, a multi-clique formulation of the data association has been proposed and solved in the BIP framework for relatively small numbers of tracks~\cite{zamir2012gmcp, dehghan2015gmmcp}, these methodologies do not scale to the large number of vehicles seen in WAMI data. SPAAM, the approach proposed in this paper, solves~\eqref{eq:OverallOptmz} approximately in a computationally tractable manner.

\begin{figure}[ht]
\center
\ifthenelse{\isdoublecol= 1}
{
\includegraphics[width=0.49\textwidth]{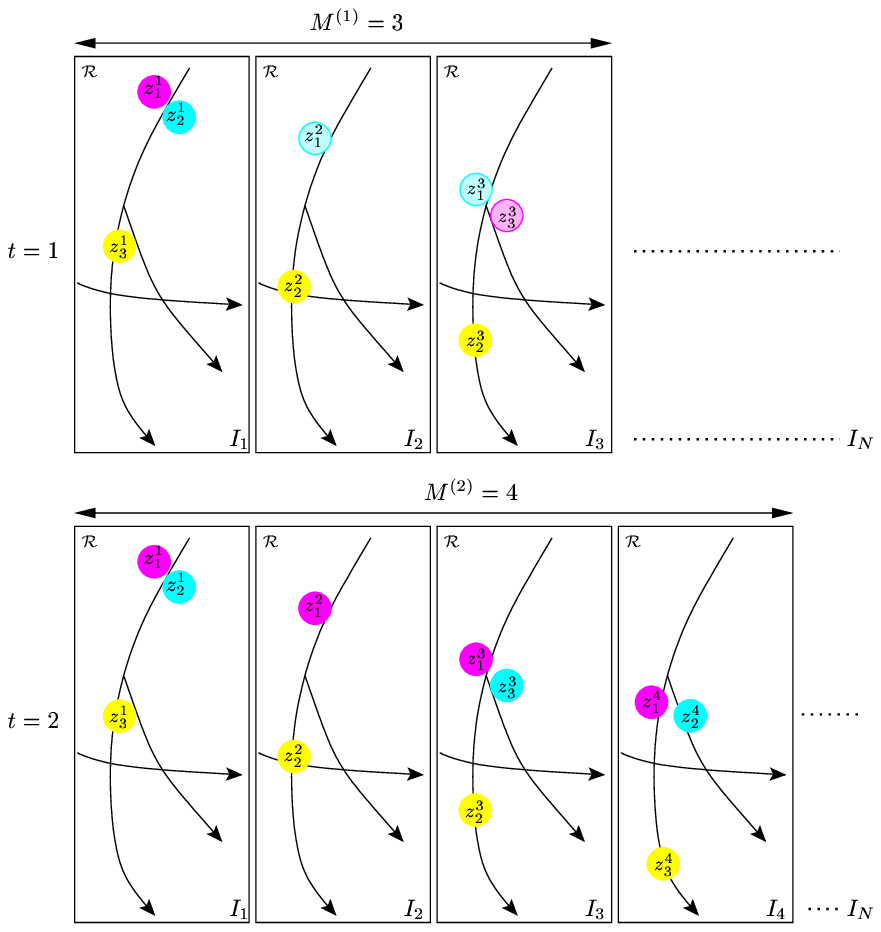}
}
{
		\resizebox{0.6\textwidth}{!}{
         \input{\FigDir/PropApproachIllustr.pstex_t}
        }
}
\caption{An example illustrating the first two iterations of SPAAM. The variable $t$ indicates the iteration number and for each $t$ the rectangular boxes arranged horizontally represent successive WAMI frames, each in the common reference coordinate system $\mathcal{R}$ with the road network overlaid (as black curves with direction of travel indicated by arrows). Detected vehicle locations associated within same track are shown in a single color hue with the saturation indicating the pairwise association confidence. The first iteration in the upper row computes associations over $\NumFrTempWn^{(1)} = 3$ frame windows and hierarchically determines associations across the windows to obtain associations over the entire temporal support $\NumFrSet$ to form tracks. The tracks are then assessed to evaluate a confidence probability for each pairwise association. In the second iteration shown in the lower row, pairwise associations are stochastically retained with a probability matching the estimated confidence probability and dissolved otherwise. The hierarchical process for estimating associations is then repeated but using a $\NumFrTempWn^{(2)} = 4$ frame temporal window. The process continues for further iterations (not shown).}
         \label{Fig:PropApproachIllustr}
\end{figure}

Operating iteratively and hierarchically, SPAAM estimates associations between adjacent vehicle detection nodes in the trellis graph $\gr$ to form the estimated vehicle tracks. The $t^{\text{th}}$ iteration first estimates associations over non-overlapping windows of $\NumFrTempWn^{(t)}$ frames each and then hierarchically computes associations across windows. The window size $\NumFrTempWn^{(t)}$ is progressively enlarged with successive iterations. To keep the complexity from rapidly growing with the increase in window size in successive iterations, only a subset of stochastically determined associations are revisited at each iteration after the first iteration. Specifically, estimated tracks at each iteration are assessed to evaluate a confidence probability for each postulated pairwise association between vehicle detection nodes and, prior to the next iteration, pairwise associations are stochastically retained with a probability matching the estimated confidence probability and disassociated otherwise. The next iteration then only considers prospective associations for the vehicle detection nodes that were stochastically disassociated while using the larger temporal context and maintaining the associations that were stochastically retained. By this process, SPAAM limits less-useful re-evaluations of association decisions that already have a high confidence and focuses available computation primarily on determining remaining associations with the benefit of a larger temporal context. A road network, accurately co-registered with the WAMI frames in  $\Frames$, provides useful spatial context for SPAAM allowing prospective associations to further be limited to road-reachable pairs of vehicle detection nodes and allowing track costs that exploit knowledge of the road topology and direction of travel. An example shown in Fig.~\ref{Fig:PropApproachIllustr} illustrates the first two iterations of SPAAM where the detection nodes are shown overlaid on a sequence of co-registered frames to establish the spatial and temporal context.

\begin{figure*}[ht]
\center
\includegraphics[width=\textwidth]{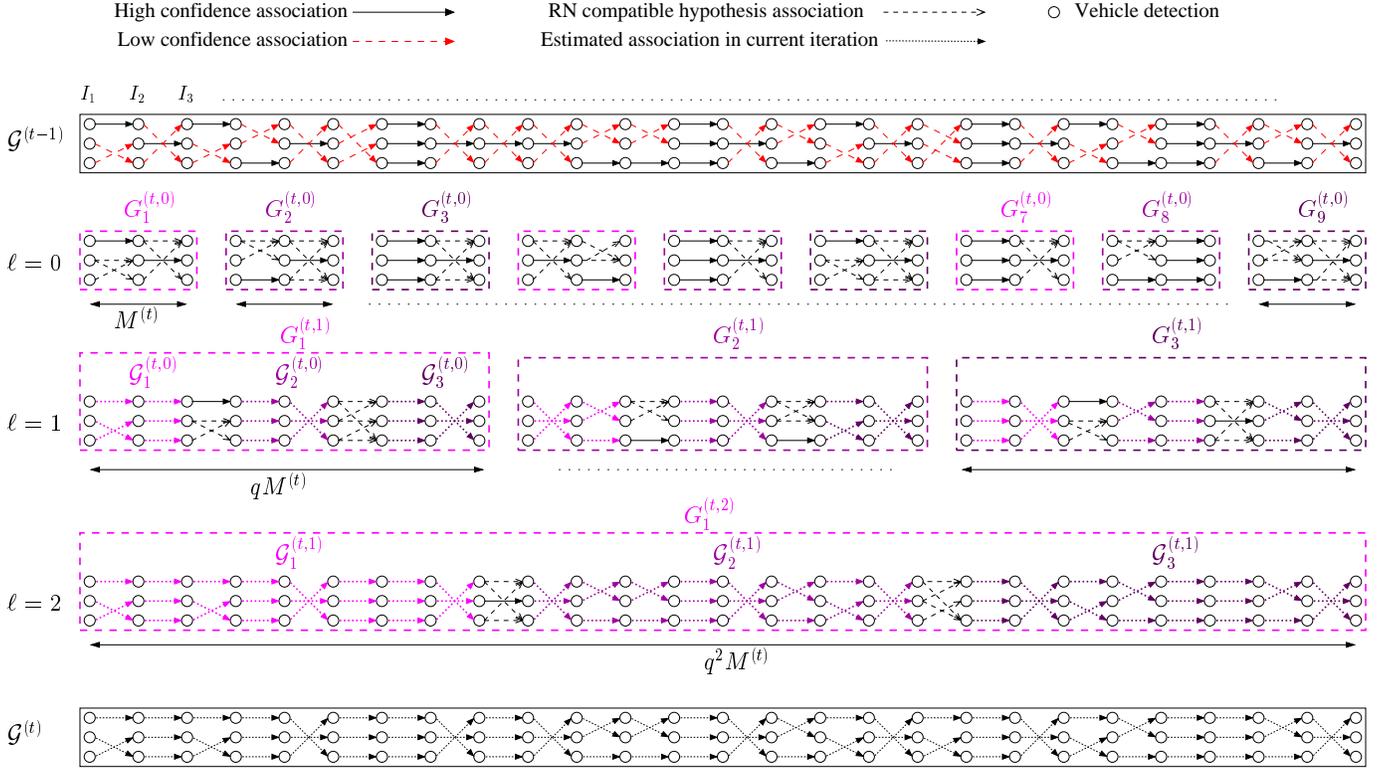}
         \caption{Hierarchical estimation of tracks in the $\nth{t}$ iteration of SPAAM. At the start of the $\nth{t}$ iteration, a confidence probability is estimated for every pairwise association (edge) in $\sgr{(\Itr-1)}{}$. Associations from  $\sgr{(\Itr-1)}{}$ are retained or dissolved based on the estimated confidence probabilities. Remaining associations are then hierarchically estimated in a common framework aimed at minimizing~\eqref{eq:OverallOptmz}. At the $\Lev=\nth{0}$ level of the hierarchy, the trellis graph is partitioned into windows of length $\NumFrTempWn^{(\Itr)}$ and for each  trellis graph $\gr^{(\Itr,\Lev)}_{w}$ over the $\nth{w}$ temporal window associations are estimated with the objective of minimizing~\eqref{eq:OverallOptmz}. Subsequent levels of the hierarchy ($l > 0$) estimate associations for temporal windows formed by groups of $q$ adjacent windows from hierarchy level $(l-1)$ using the same methodology: preserving retained associations and estimating remaining associations with the objective of minimizing~\eqref{eq:OverallOptmz}. Upon completion of the hierarchical estimation process for the $\nth{t}$ iteration of SPAAM, the set of retained and estimated associations define the edge induced subgraph $\sgr{(\Itr)}{}$ that summarizes the estimated tracks and forms the input for the next iteration.}
         \label{Fig:HierarchIllust}
\end{figure*}

Having summarized SPAAM at a high-level in the preceding paragraph, we next describe in detail the steps involved in the $\nth{t}$ iteration, relying on the trellis graph formulation developed in Section~\ref{sec:ProbFormulation}. To facilitate explanation, we use an illustrative example shown in Fig.~\ref{Fig:HierarchIllust} that also uses the trellis graph representation. At the start of $\nth{t}$ iteration, an estimated set of tracks is available from the $\nth{(t-1)}$ iteration as an edge induced subgraph $\sgr{(\Itr-1)}{}$ of the graph $\gr$. For each edge $\Edgi{i}{k_{i}}{k_{i+1}} \in E\lc\sgr{(\Itr-1)}{}\rc$,  we estimate a confidence probability $\pwc{i}{k_{i}}{k_{i+1}}$ for the pairwise association between nodes $\VDN{i}{k_i}$ and $\VDN{i+1}{k_{i+1}}$ on a path $\Tk_k \in \Pths {\sgr{(\Itr-1)}{}}$. We consider two alternatives for estimating these confidence probabilities that are detailed in Appendix~\ref{sec:associconfestim}. With the estimated probabilities we have a version of  $\sgr{(\Itr-1)}{}$ where each edge is annotated with a confidence probability for the corresponding association, as indicated in the top row in the example of Fig.~\ref{Fig:HierarchIllust}. Based on $\pwc{i}{k_{i}}{k_{i+1}}$, the edge ($\equiv$ pairwise association) $\Edgi{i}{k_{i}}{k_{i+1}}$ is stochastically retained or discarded by updating the adjacency set $\Adj^{(\Itr,0)} \lc \VDN{i}{k_i} \rc$ for the node $\VDN{i}{k_i}$ for the $\nth{t}$ iteration. Specifically independent Bernoulli random variables $\Rni{i}{k_{i}}{}$ are generated: with  $\Rni{i}{k_{i}}{}$ taking a value of $1$ with probability of $\pwc{i}{k_{i}}{k_{i+1}}$ and $0$ otherwise, and the set of disassociated nodes at the next stage of the trellis and the adjacency set are, respectively, updated as
\begin{align}
& \DisN^{(\Itr,0)}_{i+1} =  \lk \VDN{i+1}{k_{i+1}} | \Edgi{i}{k_{i}}{k_{i+1}} \in E\lc\sgr{(\Itr-1)}{}\rc \, \mbox{and} \; \Rni{i}{k_{i}}{} = 0 \;  \rk ,
\label{eq:DisNodesSetNxtStgItrLevel0} \\
& \Adj^{(\Itr,0)} \lc \VDN{i}{k_i} \rc = 
\begin{cases}
\VDN{i+1}{k_{i+1}}  &\text{if } \Rni{i}{k_{i}}{} = 1,\\
\mathcal{C} \lc \VDN{i}{k_i} \rc  &\text{if } \Rni{i}{k_{i}}{} = 0, \\
\end{cases}
\label{eq:NeighbourSetItrLevel0}
\end{align}
where $\mathcal{C} \lc \VDN{i}{k_i} \rc$ is a set of disassociated nodes in the $\nth{(i+1)}$ frame that can be reached from $\VDN{i}{k_i}$ given the road network, defined as
\ifthenelse{\isdoublecol= 1}
{
\begin{align}
   & \mathcal{C} \lc \VDN{i}{j} \rc = \nonumber \\
   & \begin{cases}
     \lk\VDN{i+1}{0},  \VDN{i+1}{1}, \ldots \VDN{i+1}{D_{i+1}}\rk & j =0, \\
     \lk \VDN{i+1}{0} \rk \bigcup \\ \lk\VDN{i+1}{j} | \vartheta \lc \VDN{i}{j}, \VDN{i+1}{j}\rc \leq \tau  \, \mbox{and} \,  \VDN{i+1}{j} \in \DisN^{(\Itr,0)}_{i+1} \rk & \mbox{otherwise,} \\
   \end{cases}
  \label{eq:RNAccessibleAdjSetDefn}
\end{align}
}
{
\begin{align}
   \mathcal{C} \lc \VDN{i}{j} \rc = 
   \begin{cases}
     \lk\VDN{i+1}{0},  \VDN{i+1}{1}, \ldots \VDN{i+1}{D_{i+1}}\rk & j =0, \\
     \lk \VDN{i+1}{0} \rk \bigcup  \lk\VDN{i+1}{j} | \vartheta \lc \VDN{i}{j}, \VDN{i+1}{j}\rc \leq \tau  \, \mbox{and} \,  \VDN{i+1}{j} \in \DisN^{(\Itr,0)}_{i+1} \rk & \mbox{otherwise,} \\
   \end{cases}
  \label{eq:RNAccessibleAdjSetDefn}
\end{align}
}
where $\vartheta\lc u_1,u_2 \rc$ is the minimum distance of travel on the road network from $u_1$ to $u_2$, and $\tau$ is a threshold that is determined based on the maximum distance a vehicle can travel between successive WAMI frames. Note that, the minimum distance of travel $\vartheta\lc u_1,u_2 \rc$ is determined by the shortest route on the road network between the two locations $u_1$ and $u_2$, which is estimated in our current implementation\footnote{Note that, the minimum distance of travel between every two points on the road network can be precomputed and stored in a look-up table for efficiency (with locations suitably quantized).} by Dijkstra's shortest path algorithm~\cite{NetworkRoutingAlgorithms:Medhi:2007}.

\begin{figure*}[ht]
\includegraphics[width=\textwidth]{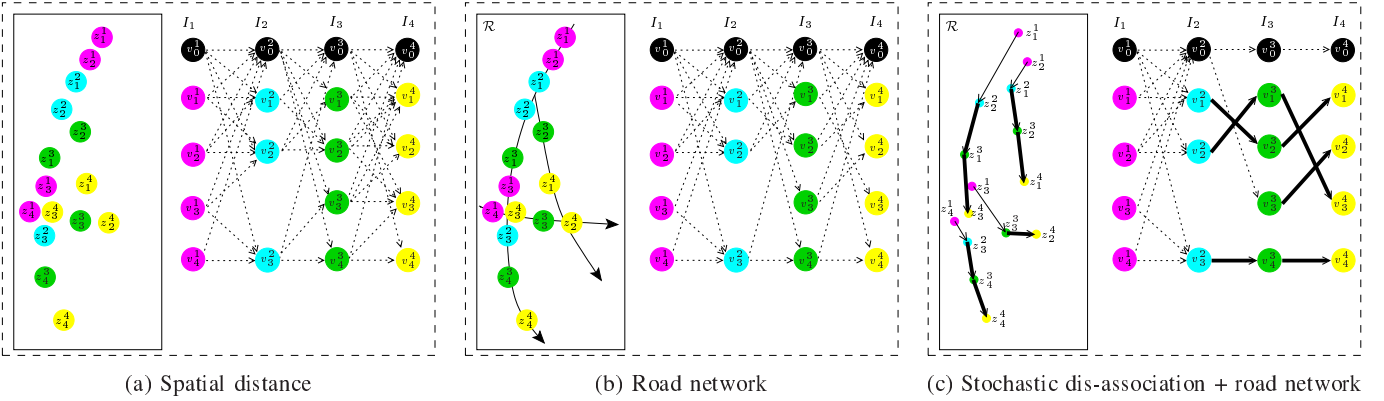}
         \caption{An example comparing alternative strategies for generating association hypotheses for a four frame WAMI 
sequence $\frm_1$, $\frm_2$, $\frm_3$, and $\frm_4$ at the first hierarchy level ($\Lev=0$) in the $\nth{\Itr}$ iteration. Distinct colors identify the VDs detected in a single WAMI frame and hypothesis associations are represented by directed edges in a corresponding trellis graph. Sub-figure~(a) illustrates a situation where the adjacency set for each node and the graph $\gr$ are formed without using the road network. In this situation, as a specific example, note that there are three nodes (plus the dummy node) that are spatially close with $\VDN{3}{3}$, so the adjacency set $\Adj^{(\Itr,0)} \lc \VDN{3}{3} \rc$ determined purely based on distance contains 4 nodes. Sub-figure~(b) illustrates the situation where the adjacency set for each node and the graph $\gr$ are formed by using the aligned road network. The graph is sparser and the adjacency sets for each node are smaller than in Fig.~\ref{Fig:PrunIllustration}~(a). Specifically, for example, $\Adj^{(\Itr,0)} \lc \VDN{3}{3} \rc$ will contains only two nodes because $\VDN{4}{2}$ is the only node within a reasonable road travel distance from $\VDN{3}{3}$.  Sub-figure~~(c) illustrates that the proposed stochastic disassociation approach where the confidence of estimated pairwise associations is represented by widths of the directed lines. The proposed approach further reduces the adjacency set for each node, specifically, in this example the association between $\VDN{3}{3}$ and $\VDN{4}{2}$ is maintained from the previous iteration and therefore, $\Adj^{(\Itr,0)} \lc \VDN{3}{3} \rc$ contains only one node.}
         \label{Fig:PrunIllustration}
\end{figure*}

As a result of the aforementioned steps, the association between $\VDN{i}{k_i}$ and $\VDN{i+1}{k_{i+1}}$ estimated in the  $\nth{(t-1)}$ iteration is \emph{dis-associated}\footnote{Note that, for a track that starts with trailing dummy nodes at iteration $(t)$, we disassociate the association from the last dummy node in the trailing dummy nodes to the first non-dummy VD node of the track at iteration $(t+1)$. Similarly, for a track that ends with trailing dummy nodes at iteration $(t)$, we disassociate the association from the last non-dummy VD node to the first dummy node in the trailing dummy nodes at iteration $(t+1)$.} with probability $\lc 1-\pwc{i}{k_{i}}{k_{i+1}} \rc$. The proposed stochastic dis-association yields adjacency sets $\Adj^{(\Itr,0)} \lc . \rc$ that contain a significantly smaller number of nodes than those obtained by naively considering all nodes in the next frame or those obtained by considering the VDs that are spatially close. This is illustrated via an example in Fig.~\ref{Fig:PrunIllustration} which highlights the fact that the combined use of the aligned road network and the stochastic disassociation in SPAAM significantly reduces the number of association hypotheses to be considered compared with an approach based on spatial distances alone.

After completion of the stochastic disassociation step, the $t^{\text{th}}$ iteration of SPAAM continues with the hierarchical estimation of tracks. This process is schematically illustrated using the trellis graph representation starting in the second row of Fig.~\ref{Fig:HierarchIllust} with $l$ denoting the hierarchy level, which starts with value of $0$ in the second row and increments in subsequent rows. At the $\nth{\Lev}$ level of the hierarchy, the temporal support of $\Frames$ is partitioned into disjoint temporal windows, each having $\NumFrLev^{\Lev}\NumFrTempWn^{(\Itr)}$ temporally adjacent frames. For the  $\nth{w}$ temporal window comprising frames $(w-1)\NumFrLev^{\Lev}\NumFrTempWn^{(\Itr)}+1$ through $w\NumFrLev^{\Lev}\NumFrTempWn^{(\Itr)}$, a corresponding trellis graph $\gr^{(\Itr,\Lev)}_{w}$ is obtained using all the nodes from $\gr$ over this temporal window and links between nodes defined by the adjacency set $\Adj^{(\Itr,\Lev)}\lc a \rc$ for each node $a$ in the graph  $\gr^{(\Itr,\Lev)}_{w}$. For the $\nth{0}$ level of the hierarchy, the adjacency sets $\Adj^{(\Itr,0)}\lc a \rc$ for nodes $a \in V\lc \sgr{\lc \Itr,\Lev-1 \rc }{m} \rc$ are as defined in~\eqref{eq:NeighbourSetItrLevel0} and for the higher levels of the hierarchy ($l>0$), these sets are defined later in this section. Tracks over each  temporal window $w$ are then estimated by solving the optimization in~\eqref{eq:OverallOptmz} over the temporal window, i.e., by determining the optimal edge induced sub-graph $\sgr{(\Itr,\Lev)}{w}$ from $\gr^{(\Itr,\Lev)}_{w}$ as
\begin{equation}
\sgr{(\Itr,\Lev)}{w} = \arg \min\limits_{\sgr{}{} \in \mathcal{F}\lc \gr^{(\Itr,\Lev)}_{w} \rc} \psi \lc \sgr{}{} \rc,
\label{eq:AnyLevelGraphEstm}
\end{equation}
where $\psi(.)$ is as defined in~\eqref{eq:Psi}. As shown in Fig.~\ref{Fig:HierarchIllust}, $w$ ranges from $1$ through $\NumFrSet/\lc \NumFrLev^{\Lev} \NumFrTempWn^{(\Itr)} \rc$, and the temporal extent of the graph $\gr^{(\Itr,\Lev)}_{w}$ spans $s\lc \gr^{(\Itr,\Lev)}_{w} \rc = (w-1)\NumFrLev^{\Lev} \NumFrTempWn^{(\Itr)}+1$ through $e\lc \gr^{(\Itr,\Lev)}_{w} \rc = w\NumFrLev^{\Lev} \NumFrTempWn^{(\Itr)}$. The adjacency sets for all nodes $a \in \gr$ are then updated for the next level of the hierarchy  as we show in Fig.~\ref{Fig:HierarchIllust}; within each temporal segment used at the current hierarchy level estimated edges are maintained, whereas across segments the results from the stochastic disassociation are used. Specifically, if $i$ is a positive multiple of $\NumFrLev^{\Lev}\NumFrTempWn^{(\Itr)}$, we set $\Adj^{(\Itr,\Lev+1)} \lc \VDN{i}{a} \rc = \Adj^{(\Itr,0)} \lc \VDN{i}{a} \rc$. Otherwise, we set $\Adj^{(\Itr,\Lev+1)} \lc \VDN{i}{a} \rc$ as the singleton set $\lk \VDN{i+1}{b} \rk$ where $\VDN{i+1}{b}$ is the node at frame $(i+1)$ estimated to associate with node $a$ in the corresponding temporal segment at level $\Lev$ of the hierarchy; formally $\Edgi{i}{a}{b} \in E\lc\sgr{\lc \Itr,\Lev \rc }{m}\rc$ where $m = \floor{i/\lc \NumFrLev^{\Lev} \NumFrTempWn^{(\Itr)} \rc}$. 

The edge induced sub-graph $\sgr{(\Itr,\Lev_{max})}{1}$ estimated at the final final hierarchy level $\Lev^{(\Itr)}_{max} = \log \NumFrSet/\lc \NumFrTempWn^{(\Itr)} \log \NumFrLev \rc$ becomes the estimated edge induced subgraph $\sgr{(\Itr)}{}$ at the end of the $t^{\text{th}}$ iteration defining the estimates of the optimal paths and associations upon completion of the $t^{\text{th}}$ iteration. The edge induced subgraph $\sgr{(\Itr)}{}$ then forms the starting point for the $(t+1)^{\text{th}}$ iteration. The overall SPAAM algorithm is summarized\footnote{The decomposed BIP problem (line 23 of Algorithm) can be solved using a variety of integer linear programming solvers such as~\cite{GurobiOptimization,GLPKOptimization,SCIPOptimization,CPLEXOptimization,CbcOptimization}.} in Algorithm~\ref{alg:StochPrgsvAlg}.

\let\oldnl\nl
\newcommand{\nonl}{\renewcommand{\nl}{\let\nl\oldnl}}
\begin{algorithm}[t]
\small
\SetAlgoLined
\SetKwInOut{Input}{Input}  
\SetKwInOut{Output}{Output} 
\Input{Vehicle detections $\VDS$ for $\NumFrSet$ WAMI frames}
\Output{Vehicle tracks $\hat{\Tks}$}
\SetKwComment{Comment}{}{}
$\NumFrTempWn^{(1)} \leftarrow \NumFrTempWn$; $V\lc \sgr{(0)}{} \rc \leftarrow \bigcup_{i=1}^{\NumFrSet} \Vrs_i$; Initialize $\sgr{(0)}{}$\; 
\For {$\Itr$ = 1 to max iter} {
$\Lev^{(\Itr)}_{max} \leftarrow \frac{\log \NumFrSet/\NumFrTempWn^{(\Itr)}}{\log \NumFrLev}$\;
{
Estimate pairwise association confidence $\pwc{i}{k_{i}}{k_{i+1}},\; \forall \Edgi{i}{k_{i}}{k_{i+1}} \in E\lc\sgr{(\Itr-1)}{}\rc$ (Appendix~\ref{sec:associconfestim})\;
Update adjacency set $\Adj^{(\Itr,0)} \lc \VDN{i}{k_i} \rc,\; \forall \VDN{i}{k_i} \in V \lc \sgr{(\Itr-1)}{} \rc $ using~\eqref{eq:NeighbourSetItrLevel0}\;
}
\For {$\Lev$ = 0 to $\Lev^{(\Itr)}_{max}$} {
\For {$w$ = 1 to $\frac{\NumFrSet}{\NumFrLev^{\Lev} \NumFrTempWn^{(\Itr)}}$} {
$s^{(\Itr,\Lev)}_w \leftarrow (w-1)\NumFrLev^{\Lev} \NumFrTempWn^{(\Itr)}+1$; $e^{(\Itr,\Lev)}_w \leftarrow w\NumFrLev^{\Lev} \NumFrTempWn^{(\Itr)}$\;
\Comment{\textbf{\underline{Obtain $V\lc \gr^{(\Itr,\Lev)}_{w} \rc$ and $E\lc \gr^{(\Itr,\Lev)}_{w} \rc$}}}
\Comment{\textbf{\underline{to construct $\gr^{(\Itr,\Lev)}_{w}$:}}}
$V\lc \gr^{(\Itr,\Lev)}_{w} \rc \leftarrow \bigcup_{i=s^{(\Itr,\Lev)}_w}^{e^{(\Itr,\Lev)}_w} \Vrs_i$\;
\If{$\Lev \neq 0$}{
	\For {$i$ = $s^{(\Itr,\Lev)}_w$ to $e^{(\Itr,\Lev)}_w$} {
	\For {$a$ = $1$ to $D_{i+1}$} {
		\eIf{$i \mod \NumFrLev^{\Lev-1} \NumFrTempWn^{(\Itr)} = 0$}{
			$\Adj^{(\Itr,\Lev)} \lc \VDN{i}{a} \rc \leftarrow \Adj^{(\Itr,0)} \lc \VDN{i}{a} \rc$\;
			}{
			$m\leftarrow \floor{i/\lc \NumFrLev^{\Lev-1} \NumFrTempWn^{(\Itr)} \rc}$\;
			$\Adj^{(\Itr,\Lev)} \lc \VDN{i}{a} \rc \leftarrow \{ \VDN{i+1}{b} | \Edgi{i}{a}{b} \in E\lc\sgr{\lc \Itr,\Lev-1 \rc }{m}\rc \}$\;
			}
	}
	}
}
$E\lc \gr^{(\Itr,\Lev)}_{w} \rc \leftarrow \lk \Edgi{i}{a}{b} \left| \VDN{i+1}{b}  \in \Adj^{(\Itr,\Lev)} \lc \VDN{i}{a} \rc,  \VDN{i}{a},\VDN{i+1}{b} \in V\lc \gr^{(\Itr,\Lev)}_{w} \rc \right. \rk$\;
Estimate $ \sgr{(\Itr,\Lev)}{w}$ from $\gr^{(\Itr,\Lev)}_{w}$ using~\eqref{eq:AnyLevelGraphEstm}\;
}
}
$ \sgr{(\Itr)}{} \leftarrow  \sgr{(\Itr,\Lev^{(\Itr)}_{max})}{1}$; $\NumFrTempWn^{(\Itr+1)} \leftarrow \NumFrTempWn^{(\Itr)} + 1$\;
}
$\hat{\Tks} \leftarrow \Pths {\sgr{(\Itr)}{}}$\;
\caption{Stochastic progressive association across multiple frames (SPAAM) \label{alg:StochPrgsvAlg}}
\end{algorithm}

\section{Experimental results} \label{sec:results}

\begin{figure*}[ht!]
\includegraphics[width=\textwidth]{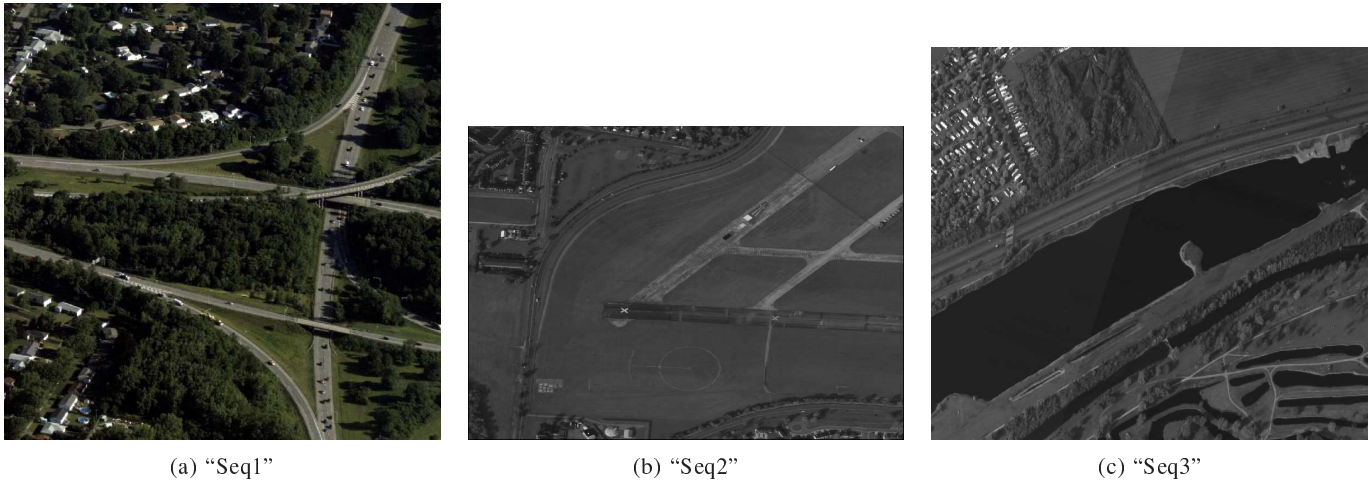}
        \caption{Test sequences used in our evaluation. Each test sequence contains 60 frames and we show the first frame from ``Seq1'', ``Seq2'', and ``Seq3'', in (a), (b), and (c) respectively. All roads in ``Seq1'' and ``Seq3'' are one-way roads, while roads in ``Seq2'' are two-way roads.}
\label{fig:TestSequences}
\end{figure*}

To evaluate tracking performance, we use three test sequences with distinct characteristics chosen from two WAMI data sets: (a) CorvusEye dataset that is recorded using the CorvusEye 1500 Wide-Area Airborne System\cite{CorvusEye} over the Rochester, NY region, and (b) Wright-Patterson Air Force Base (WPAFB) 2009 dataset~\cite{WPAFB}, which was recorded over the WPAFB, OH region. Fig. \ref{fig:TestSequences}  depicts a frame from each sequence of the three sequences that we label as ``Seq1'', ``Seq2'' and ``Seq3''.  ``Seq1'' uses a region from the CorvusEye dataset that contains forked one-way roads with different directions and also several occluders (bridges, trees, etc.). ``Seq2'' and ``Seq3'' correspond to two different regions from the WPAFB dataset. Both ``Seq2'' and ``Seq3'' cover regions that are free from occluders, however,  ``Seq2'' contains only two-way roads,  ``Seq3'' has only one-way roads. All test sequences comprise $60$ ground truth labeled frames\footnote{The detection locations that form the inputs to the algorithm, along with the aligned road network, and labeled ground truth tracks of the three sequences are available at \url{http://www.ece.rochester.edu/~gsharma/VisDataAnalGeoSpat/}}. 
For the vector road map, we use OpenStreetMap (OSM)~\cite{OSM}, which provides each road in the road network in a vector format along with properties of each road such as type (highway, residential, etc), one or two-way traffic, number of lanes, etc. We use the method in~\cite{RoadNetRegistration:tip:2016} for co-registering the WAMI frames to the vector road map. We obtain the VDs via the background subtraction method used in~\cite{DetectTrackLarge2010}. We set $\NumFrTempWn^{(1)} = 3$ in all experiments and we use the Gurobi optimizer~\cite{GurobiOptimization} for solving~\eqref{eq:AnyLevelGraphEstm}. The Gurobi optimizer relaxes the integer restrictions in the binary constraints in~\eqref{eq:ILPConstraint1} and uses a linear-programming based branch-and-bound algorithm for obtaining a solution\footnote{More information about the algorithm used by the Gurobi optimizer can be found at http://www.gurobi.com/resources/getting-started/mip-basics}. For all experiments, the parameters in~\eqref{eq:HypsCost} are set to empirically determined values of $\sigma_m = 12.5$, $\sigma_d = 0.02$, $\sigma_{\theta} = 100$, $\sigma_g = 2.8$, and $W=2$.

Several methods are benchmarked. Two instances of the proposed method are considered: \textbf{SPAAM-M} and \textbf{SPAAM-EM}, which, respectively, estimate the confidence of pairwise associations using the marginalization in~\eqref{eq:PrWsAssocConfMrg} or the EM approach in~\eqref{eq:PrWsAssocConfEM}. We compare the proposed SPAAM instances with the multi-data association, iterative conditional modes like method (\textbf{{\IGT}}) of~\cite{MDAHigherOrderModels:collins:2012}. The {\IGT} method does not require per association independent additive costs for tracks and is used with the same cost function as SPAAM (defined in~\eqref{eq:HypsCost}) allowing for a fair comparison of the techniques without differences induced by the underlying cost function. The methods are also compared against an  online tracking method \textbf{{\OT}} proposed in~\cite{VehcDetectTrack2010} that considers a pre-aligned road network when estimating the frame-to-frame VDs associations. The tracks estimated by the online tracking method also serve as the initialization for all of the iterative methods, specifically, the different instances of SPAAM and for {{\IGT}}.

\newcommand\NameEntry[2]{%
  \multirow{#1}*{%
    \begin{varwidth}{4em}
    \centering #2%
    \end{varwidth}}}

\begin{table}[h]
\centering
\begin{tabular}{|c|c|c|c|c|c|c|c|}
\hline
Seq& Method & \textit{MT}$\uparrow$ & \textit{ML}$\downarrow$ & \textit{PT} &\textit{IDS}$\downarrow$&\textit{Frag}$\downarrow$ & \textit{MOTA}$\uparrow$\\
\hline
  \NameEntry{3}{1} & {\OT} & 73 & {11} &55& 292&384 & 0.854994\\
\cline{2-8}
  &   {\IGT}   & 81  & 15 & 43 & 164 & 211 & 0.91953  \\                                                            
\cline{2-8}
  &  SPAAM-M  & {85} & {14} & 40 & {110}& {176} & 0.946249\\   
    \cline{2-8}
  &  SPAAM-EM  & {80} & 16 & 43 & \textbf{82}& \textbf{144} & \textbf{0.959503} \\              
\hline                          
\thickhline
  \NameEntry{3}{2} & {\OT} & 94 & 2 &6& 311& 360 & 0.884093 \\
\cline{2-8}
  &   {\IGT}   & 95 & 3 & 4 &257 &290 & 0.902187 \\
  \cline{2-8}
  &  SPAAM-M  & {97} & 3 & 2 & {101}& {128} & 0.961854\\
  \cline{2-8}
  &  SPAAM-EM & {97} & 3 & 2 & \textbf{60}& \textbf{83} & \textbf{0.977017}  \\
\hline          
\thickhline
  \NameEntry{3}{3} & {\OT} & 53 & 0 & 1 & 112 & 138 & 0.932051 \\ 
\cline{2-8}
  &   {\IGT}   & 51 & 1 & 2 & 86 & 101 & 0.944871 \\                                 
\cline{2-8}
  &  SPAAM-M  & 54 & 0 & 0 & 40 & 59 & 0.974878\\                         
  \cline{2-8}
  &  SPAAM-EM  & {54} & 0 & 0 & \textbf{27}& \textbf{39} &  \textbf{0.983806}\\ 
\hline
\thickhline
\end{tabular}
\caption{Comparison of tracking performance metrics for four different tracking algorithms on the three different WAMI sequences (datasets). The metrics include mostly tracked (MT), mostly lost (ML), partly tracked (PT), ID switches (IDS), fragments (Frag), and multiple object tracking object accuracy (MOTA). See text for definitions of the metrics: the annotations $\uparrow$ and $\downarrow$ next to a metric in the table indicate whether a higher or lower value is better, respectively. The four methods compared include: the {\OT} method in~\cite{VehcDetectTrack2010}, the two alternative proposed SPAAM instances (SPAAM-M and SPAAM-EM) and the {\IGT} method in~\cite{MDAHigherOrderModels:collins:2012}. While both the  {\IGT} method and the proposed SPAAM instances  (SPAAM-M and SPAAM-EM) improve upon the {\OT} method, the performance metrics for the  proposed SPAAM instances  (SPAAM-M and SPAAM-EM) are significantly better than those for the {\IGT} method, despite the fact that both use the same cost function and initialization. 
}
\label{Tab:HigherOrderModelsPropComprsn}
\end{table}

We quantify tracking performance with reference to  labeled ground-truth tracks by the widely adopted measures defined in~\cite{PartiallyOccludedHumansTracking:2006:wu} which are (1) the total number of ID switches (\textit{IDS}$\downarrow$)\footnote{$\uparrow$ and $\downarrow$ indicate that higher and lower is better, respectively.} for estimated tracks compared to the ground truth, (2) the number of mostly tracked (\textit{MT}$\uparrow$) vehicles, i.e., the vehicles for which estimated tracks include over 80\% of the detections in the ground truth, (3) the number of mostly lost (\textit{ML}$\downarrow$), i.e., the vehicles for which estimated tracks include under  20\% of the detections in the ground truth, (4) the number of partially tracked (\textit{PT}$\downarrow$), i.e., the vehicles for which estimated tracks include below 80\% and over 20\% of the detections in the ground truth, and (5) the number of track fragments (\textit{Frag}$\downarrow$), where a fragment is defined as a part of vehicle track with length less than 80\% of the length of the corresponding ground truth track. In addition to these measures, we report the Multiple Object Tracking Accuracy (\textit{MOTA}$\uparrow$) defined in~\cite{MOTEval:Kasturi:2009} (with $c_s = 1$). Table~\ref{Tab:HigherOrderModelsPropComprsn} enumerates these metrics for the tracks estimated for the three WAMI datasets using the four methods: {\OT} method in~\cite{VehcDetectTrack2010}, the two proposed  instances of SPAAM (SPAAM-M and SPAAM-EM), and the {\IGT} method~\cite{MDAHigherOrderModels:collins:2012}. The results in the table indicate that the proposed SPAAM instances and the {\IGT} method both improve upon the {\OT} method that serves as the initialization for both these methods. However, the improvement offered by the proposed SPAAM instances is much larger than that for the {\IGT} method. Overall, both the proposed  SPAAM instances (SPAAM-M and SPAAM-EM) perform significantly better than the {\IGT} method.

Table~\ref{Tab:MDAandPropExecTimeComprsn} lists the execution times for the two proposed SPAAM instances (SPAAM-M and SPAAM-EM) and for the {\IGT} method along with the corresponding number of iterations. From the table it can be seen that the proposed SPAAM instances have much lower execution times. Thus the proposed SPAAM approach offers performance gains in both tracking accuracy and in execution time.

\begin{table}
\begin{center}
\begin{tabular}{|c|c|c|c|c|c|c|}
\hline
 \NameEntry{2}{Seq} &  \multicolumn{2}{ c  }{SPAAM-EM} & \multicolumn{2}{| c }{SPAAM-M} & \multicolumn{2}{ |c|  }{{\IGT}}  \\
\cline{2-7}
   &  Iters & Time & Iters & Time & Iters & Time \\
\hline
  1 & 6 & 0.1490 & 6 & 0.1479 & 23 & 26.8306\\
\hline     
  2 & 6 & 0.2003 & 6 & 0.2142 & 13 & 6.1653 \\  
\hline 
  3 & 6 & 0.0833 & 6 & 0.0235 & 22 & 3.4917 \\               
\hline                      
\end{tabular}
\end{center}
\caption{Execution time (in hours) for the two proposed SPAAM instances (SPAAM-M and SPAAM-EM) and for the {\IGT} method along with the number of corresponding iterations (Iters). All algorithms are executed on a computer with an Intel\textsuperscript{\textregistered} Core\TM\; i7-6700HQ CPU with 8 cores, operating at 2.60 GHz, having 12 GB of main memory, and running the Ubuntu 16.04 LTS operating system. The proposed SPAAM instances have much lower execution times.}

\label{Tab:MDAandPropExecTimeComprsn}
\end{table}

\section{Discussion}
\label{sec:discussion}

\begin{figure}[h]
\center
\includegraphics[width=0.48\textwidth]{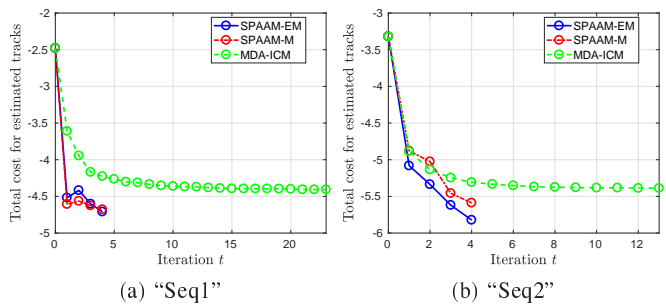}
         \caption{The total cost of the estimated tracks  with progression of iterations for the proposed SPAAM instances (SPAAM-M and SPAAM-EM) and for the {\IGT} method in~\cite{MDAHigherOrderModels:collins:2012} for: (a) ``Seq1'' and (b) ``Seq2''. The methods use a common cost function and a common initialization. The {\IGT} technique demonstrates monotonic improvement of the cost with iterations, whereas SPAAM-M and SPAAM-EM do not exhibit monotonic improvement. However, with the progression of iterations, SPAAM-M and SPAAM-EM achieve a lower total cost than {\IGT}, which results in a better set of estimated tracks.}
         \label{Fig:AccuracyIteration}
\end{figure}

The performance of the {\OT} method is hampered by the fact that it minimizes a cost function that is a sum of independent costs for pairwise associations between adjacent frames. In situations like WAMI vehicle tracking, where only vehicle detection locations are available, such cost functions are fundamentally limited because they do not allow the use of a larger temporal context for assessing plausibility of (postulated) vehicle movements. As a result, although {\OT} method minimizes its cost function via the estimation of associations from frame to frame in temporal sequence, the limitations of the cost function make it prone to tracking errors. The {\IGT} and proposed SPAAM methods, use a cost function that is better suited for WAMI vehicle tracking by using a larger temporal context to assess vehicular movements. This allows them to improve upon the performance of the {\OT} method, even though the methods are not able to guarantee that a global minimum of the cost function is achieved. To further understand the differences between the {\IGT}  and the  proposed SPAAM approach, it is instructive to compare the evolution of the (common) cost function for these methods with the progression of iterations. Figure~\ref{Fig:AccuracyIteration} shows the total cost for tracks estimated using the proposed SPAAM instances (SPAAM-M and SPAAM-EM) and the {\IGT} method as a function of the iteration number for these methods. In each step of its iterative process, the {\IGT} method re-estimates all of the associations across temporally adjacent frames to improve the global cost, while keeping all other associations fixed. This approach guarantees that the iterations improve monotonically assuring convergence of the {\IGT} approach to a local minimum, a property it inherits from the iterative conditional modes (ICM)~\cite{ICM_StatisticalAnalysisDirtyPictures:Besag:1986} approach on which it is based. While the approach ensures computational tractability, it cannot avoid getting trapped in poor local minima in the vicinity of the initialization. On the other hand, via the stochastic disassociation and progressive enlargement of temporal context, the SPAAM approach revisits (predominantly) low confidence associations collectively with the benefit of additional information available from the enlarged context. Thus, even though SPAAM does not guarantee monotonic convergence, it is better at avoiding local minima and, as demonstrated in the results, typically offers a significant improvement in tracking accuracy.

\begin{figure}[h]
\center
\includegraphics[width=0.48\textwidth]{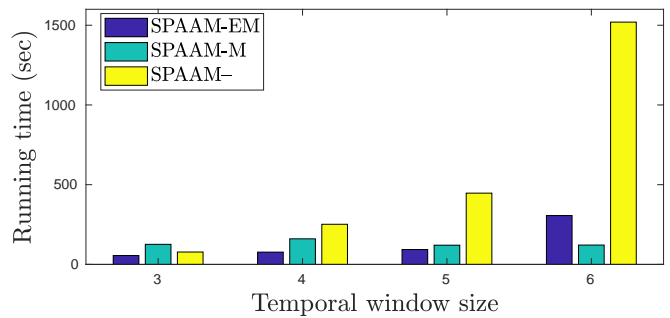}
         \caption{Execution time (in seconds) for different temporal window sizes for SPAAM-M, SPAAM-EM, and the de-featured variant SPAAM-- that estimates all associations afresh. ``Seq1'' was used for obtaining these execution times. The figure illustrates the key advantages of the SPAAM-M and SPAAM-EM approaches: with progressive enlargement of the temporal window size the computation increases only modestly, whereas the execution time increases exponentially with increasing window size for SPAAM--. The hardware configuration used for these experiments is identical to that specified in Table~\ref{Tab:MDAandPropExecTimeComprsn}.}
         \label{Fig:ProgrsvTime}
\end{figure}

To highlight the benefit of stochastic dis-association and the progressive enlargement of the temporal window sizes in SPAAM, we consider \textbf{SPAAM--}, a deliberately de-featured version of proposed approach that estimates all of the associations in single iteration using a fixed temporal window size. The SPAAM-- approach is obtained in Algorithm~\ref{alg:StochPrgsvAlg} by setting \textit{max iterations} = 1 and $E\lc \sgr{(0)}{} \rc = \phi$. Fig.~\ref{Fig:ProgrsvTime} compares the execution times required for estimation of tracks for ``Seq1'' using the proposed SPAAM-M, SPAAM-EM, and the de-featured variant SPAAM-- with different temporal window sizes\footnote{For SPAAM-M and SPAAM-EM these are the corresponding iteration execution time.}. The figure shows that the execution time for the SPAAM-- increases exponentially with the increase in the size of the temporal window used for estimating the associations. On the other hand, the SPAAM-M and SPAAM-EM approaches experience only a small increase in the execution time with progressive enlargement of the temporal window sizes because, with high probability, they maintain pairwise associations already estimated in previous iteration with high confidence in the previous iteration and revisit only the low confidence associations. Thus, the number of association  hypotheses has a limited increase with the enlargement of the temporal window size, which is reflected as a corresponding small increase in the execution time seen in the figure.

Finally, we note that although this paper focused specifically on the problem of vehicle tracking in WAMI, the SPAAM approach is more general and may offer a useful framework for other trellis graph association problems where: (a) it is beneficial to use cost functions that use a larger temporal context that disallows a pairwise independent cost decomposition and (b)  one can assess confidence of associations. Exploration of such alternatives is beyond the scope of the present paper but would make interesting future work.

\section{Conclusion}
\label{sec:concl}

The stochastic progressive association across multiple frames (SPAAM) framework proposed in this paper provides a novel, computationally-efficient, iterative, and accurate approach for estimating vehicular tracks in WAMI. Specifically, SPAAM allows the use of effective cost functions that assess the regularity of vehicular motion over a  multi-frame window while simultaneously maintaining computational tractability. These dual objectives are achieved  via a strategy of estimating tracks hierarchically over progressively increasing window lengths. As the window is enlarged, high confidence estimated associations are predominantly maintained and low confidence associations are predominantly dissolved, and tracks are re-estimated with the larger temporal context. The stochastic approach to maintaining/dissolving associations reduces the computation required for larger windows while allowing the tracks to be improved through the incorporation of additional information provided by the larger context. The combination of SPAAM with information provided by a co-registered road network allows the proposed method to tackle the challenges of scale and limited spatial detail in WAMI tracking. Results obtained over three test sequences that represent different tracking scenarios show a significant performance improvement for the proposed approach compared with three state of the art alternative methods.

\section*{Acknowledgment}  \label{sec:ack}
We thank Bernard Brower of Harris Corporation for making available the
CorvusEye~\cite{CorvusEye} WAMI datasets used in this research.

\appendix

\section{Estimation of confidence for predicted pairwise associations} \label{sec:associconfestim}

Confidence for pairwise associations is estimated via two alternative probabilistic modeling approaches. The first, uses an energy model to transform costs for tracks into probabilities, followed by approximate marginalization to estimate the probability of an association. The second approach uses a two component mixture model for motion features. Parameters for the model are obtained using semi-supervised expectation maximization (EM) and the association probability is estimated as the posterior probability of the corresponding mixture component.

\subsection*{Energy model based marginalization for pairwise association confidence estimation}

An energy model~\cite{PRML:Bishop:2006} can be used to coherently extend the costs we already use into a corresponding probabilistic model. Specifically, the probability of an edge induced sub-graph $\sgr{}{} \in \mathcal{F}\lc \gr \rc$  can be modeled as $P\lc \sgr{}{} \rc = \frac{1}{Z_{\gr}} e^{-\psi \lc \sgr{}{} \rc}$, where $Z_{\gr} = \sum\limits_{\substack{\sgr{}{} \in \mathcal{F}\lc \gr \rc}} e^{-\psi\lc \sgr{}{} \rc}$ is partition function that ensures probabilities sum to $1$. The minimization in~\eqref{eq:OverallOptmz} is then equivalent to a maximum a posterior (MAP), or minimum energy, estimate. With the probabilistic model, the confidence $\pwcX{i}{a}{b}$ of a pairwise association that associates $\VDN{i}{a}$ with $\VDN{i+1}{b}$ can be estimated as the marginal probability of inclusion of the edge $\Edgi{i}{a}{b}$ in the sub-graph. Formally, $\pwcX{i}{a}{b} = \sum\limits_{\substack{\sgr{}{} \in \mathcal{F}\lc \gr \rc,\\ \Edgi{i}{a}{b} \in E\lc \sgr{}{} \rc }} P\lc \sgr{}{}\rc.$

The above marginalization is intractable because of the dependency introduced by the feasibility constraints. Therefore, we approximate $\pwcX{i}{a}{b}$ by 
\begin{equation}
\pwc{i}{a}{b} = \frac{1}{Z_a^i} \; \sum\limits_{\substack{\Tk_k \in \Pths{\gr}, \\ \VDN{i}{a}, \VDN{i+1}{b} \in \Tk_k}} e^{-C\lc \Tk_k \rc},
\label{eq:PrWsAssocConfMrg}
\end{equation}
where $Z_a^i = \sum\limits_{\VDN{i+1}{m} \in \Adj\lc \VDN{i}{a} \rc} \sum\limits_{\substack{\Tk_k \in \Pths{\gr}, \\ \VDN{i}{a}, \VDN{i+1}{m} \in \Tk_k}} e^{-C\lc \Tk_k \rc}$ is a normalization constant. The above approximation is reasonable for the following reasons that are apparent from~\eqref{eq:PrWsAssocConfMrg}. First, $\VDN{i}{a}$ must be associated with only one of the VDs in $\Adj\lc \VDN{i}{a} \rc$, and therefore, $\sum\limits_{\VDN{i+1}{m} \in \Adj\lc \VDN{i}{a} \rc} \pwc{i}{a}{m} = 1$. Second, associating   
$\VDN{i}{a}$ with $\VDN{i}{b}$ is more probable than associating with $\VDN{i}{c}$ if the sum of the costs of the tracks that contain the pairwise association between $\VDN{i}{a}$ and $\VDN{i}{b}$ is smaller than the tracks that contain the pairwise association between $\VDN{i}{a}$ and $\VDN{i}{c}$. 

\subsection*{Semi-supervised EM based mixture model}

To estimate the confidence of a pairwise association $\Edgi{i}{a}{b}$ that associates $\VDN{i}{a}$ with $\VDN{i+1}{b}$, we first estimate temporally local features $\Ftr{i}{a,b} = \lsq \Gamma_m \lc \Edgi{i}{a}{b} \rc, \Gamma_\theta \lc \Edgi{i}{a}{b} \rc, R_d \lc \Edgi{i}{a}{b} \rc, R_\theta \lc \Edgi{i}{a}{b} \rc \rsq$ for the pairwise association, then we model the distribution of $\Ftr{i}{a,b}$ as a two component Gaussian mixture model (GMM)~\cite{PRML:Bishop:2006}. One mixture component corresponds to the case when $\VDN{i}{a}$ {\em{associates}} with $\VDN{i+1}{b}$ (denoted as $\VDN{i}{a} \rightarrow \VDN{i+1}{b}$), while the other component corresponds to the case when $\VDN{i}{a}$ {\em{does not associate}} with $\VDN{i+1}{b}$. To indicate which mixture component is responsible for the pairwise association, we associate the pairwise association with a  binary latent variable $\Edgl{i}{a}{b}$, where 
\begin{equation}
\Edgl{i}{a}{b} = \begin{cases}
1  &\text{if }  \VDN{i}{a} \rightarrow \VDN{i+1}{b}, \\
0   &\text{otherwise}. \\
\end{cases}
\nonumber
\end{equation}
The confidence of the pairwise association is then approximated by the posterior probability that $\Edgl{i}{a}{b}=1$ given the computed features $\Ftr{i}{a,b}$.

Specifically, the two GMM components are parametrized by the corresponding means ${\boldsymbol \mu}_r$ and covariances ${\boldsymbol \Sigma}_r$ of the features, where $r\in\lk 0,1\rk$. Also, $\Edgl{i}{a}{b}$ is assumed to be Bernoulli with parameter $\gamma = p \lc \Edgl{i}{a}{b}=1 \rc$. Thus, the GMM is parametrized by ${\boldsymbol \theta} = \lk \gamma, {\boldsymbol \mu}_0, {\boldsymbol \mu}_1, {\boldsymbol \Sigma}_0, {\boldsymbol \Sigma}_1 \rk$. The likelihood of $\Ftr{i}{a,b}$ is
\begin{align}
p \lc \Ftr{i}{a,b} \left|{\boldsymbol \theta} \right. \rc = \sum\limits_{r=0}^{1} p \lc \Ftr{i}{a,b}, \Edgl{i}{a}{b}=r \left|{\boldsymbol \theta} \right. \rc,
\label{eq:GMM0}
\end{align}
where
\ifthenelse{\isdoublecol= 1}
{
\begin{align}
p \lc \Ftr{i}{a,b}, \Edgl{i}{a}{b}=r \left|{\boldsymbol \theta} \right. \rc &= p \lc \Edgl{i}{a}{b}=r \left| {\boldsymbol \theta} \right. \rc p \lc \Ftr{i}{a,b}\left| \Edgl{i}{a}{b}=r, {\boldsymbol \theta} \right. \rc,\nonumber\\
& = \gamma^r \lc1-\gamma \rc^{1-r}\; \mathcal{N}\lc \Ftr{i}{a,b} \left| {\boldsymbol \mu}_r, {\boldsymbol \Sigma}_r \right. \rc.
\label{eq:GMM1}
\end{align}
}
{
\begin{align}
p \lc \Ftr{i}{a,b}, \Edgl{i}{a}{b}=r \left|{\boldsymbol \theta} \right. \rc &= p \lc \Edgl{i}{a}{b}=r \left| {\boldsymbol \theta} \right. \rc p \lc \Ftr{i}{a,b}\left| \Edgl{i}{a}{b}=r, {\boldsymbol \theta} \right. \rc,\nonumber\\
& = \gamma^r \lc1-\gamma \rc^{1-r}\; \mathcal{N}\lc \Ftr{i}{a,b} \left| {\boldsymbol \mu}_r, {\boldsymbol \Sigma}_r \right. \rc.
\label{eq:GMM1}
\end{align}
}
The confidence of the pairwise association is then computed as
\begin{align}
\pwc{i}{a}{b} &= p \lc \Edgl{i}{a}{b} = 1 \left| \Ftr{i}{a,b}, \hat{{\boldsymbol\theta}} \right. \rc,\nonumber\\
&=\frac{p \lc \Ftr{i}{a,b}, \Edgl{i}{a}{b} = 1 \left|\hat{{\boldsymbol \theta}} \right. \rc}{ \sum\limits_{r=0}^{1} p \lc \Ftr{i}{a,b}, \Edgl{i}{a}{b} =r \left|\hat{{\boldsymbol \theta}} \right. \rc},
\label{eq:PrWsAssocConfEM}
\end{align}
where $\hat{{\boldsymbol \theta}} = \lk \hat{\gamma}, \hat{{\boldsymbol \mu}_0}, \hat{{\boldsymbol \mu}_1}, \hat{{\boldsymbol \Sigma}_0}, \hat{{\boldsymbol \Sigma}_1} \rk$ are the estimated parameters of the GMM. We estimate $\hat{{\boldsymbol \theta}}$ using the EM framework~\cite{EM:Dempster:1977} through a semi-supervised approach. The semi-supervised approach exploits available ground truth training data (for example from~\cite{WPAFB}). Let $\tilde{\mathcal{G}}$ be a ground truth trellis graph\footnote{Note that the ground truth trellis graph is constructed using ground truth tracks in different dataset, i.e., $V\lc \sgr{}{} \rc \neq V\lc \tilde{\mathcal{G}} \rc$.} that is constructed from the ground truth tracks such that every ground truth track represents a path in $\tilde{\mathcal{G}}$. Then, the EM framework estimates $\hat{{\boldsymbol \theta}}$ by
\ifthenelse{\isdoublecol= 1}
{
\begin{align}
\hat{{\boldsymbol \theta}} = \arg\max\limits_{{\boldsymbol \theta}} &\prod\limits_{\Edgi{i}{a}{b} \in E\lc\tilde{\mathcal{G}}\rc} p \lc \Edgi{i}{a}{b}, \Edgl{i}{a}{b} =1 \left|{\boldsymbol \theta} \right. \rc\times\nonumber\\
&\prod\limits_{\Edgi{i}{a}{b} \in E\lc\sgr{}{}\rc} p \lc \Ftr{i}{a,b} \left|{\boldsymbol \theta} \right. \rc.
\label{eq:Likhd}
\end{align}
}
{
\begin{align}
\hat{{\boldsymbol \theta}} = \arg\max\limits_{{\boldsymbol \theta}} &\prod\limits_{\Edgi{i}{a}{b} \in E\lc\tilde{\mathcal{G}}\rc} p \lc \Edgi{i}{a}{b}, \Edgl{i}{a}{b} =1 \left|{\boldsymbol \theta} \right. \rc\times\nonumber\\
&\prod\limits_{\Edgi{i}{a}{b} \in E\lc\sgr{}{}\rc} p \lc \Ftr{i}{a,b} \left|{\boldsymbol \theta} \right. \rc.
\label{eq:Likhd}
\end{align}
}
under the assumption that the pairwise associations are conditionally independent\footnote{Note that $\hat{{\boldsymbol \theta}}$ can be also estimated through an unsupervised approach, for which $E\lc\tilde{\mathcal{G}}\rc = \phi$. To identify the mixture component that is corresponds to $\VDN{i}{a} \rightarrow \VDN{i+1}{b}$ case, the estimated covariances of the two Gaussian components are compared and the component that has smaller parameter variance is selected.}.

\ifCLASSOPTIONcaptionsoff
  \newpage
\fi


\end{document}